\documentclass[12pt]{article}
\usepackage{amsthm,amsmath,amssymb,mathtools}
\usepackage{latexsym}
\usepackage{graphicx}
\usepackage{url}
\usepackage{hyperref}

\usepackage{xcolor}

\usepackage{bm}
\usepackage{comment}
\usepackage{statex}
\usepackage{diagbox}
\usepackage{multirow}
\usepackage{chngpage}
\usepackage{verbatim} 
\usepackage{booktabs} 
\usepackage{cite}

\usepackage{lineno}
\usepackage{array}
\usepackage{bbm}
\usepackage{amsmath, amsfonts, amssymb, amsthm, amsxtra}
\usepackage{mathrsfs, mathtools}
\usepackage{enumitem}
\usepackage{stmaryrd}

\usepackage{algorithm}
\usepackage{algorithmic}

\def\@themcountersep{}

\newtheorem*{theorem*}{Theorem}
\newtheorem{theorem}{Theorem}

\newtheorem*{prop*}{Proposition}
\newtheorem{rema}{Remark}
\newtheorem*{rema*}{Remark}
\newtheorem{lemma}[theorem]{Lemma}
\newtheorem{definition}{Definition}
\newtheorem*{definition*}{Definition}
\newtheorem*{coro*}{Corollary}
\newtheorem{coro}{Corollary}

\definecolor{green}{rgb}{0,0.50,0.10}

\DeclareMathOperator*{\argmin}{arg\,min}

\newcolumntype{L}{>{\centering\arraybackslash}m{2.2cm}}

\def\0{\boldsymbol 0}
\def\1{\boldsymbol 1}
\def\2{\boldsymbol 2}
\def\3{\boldsymbol 3}
\def\4{\boldsymbol 4}
\def\5{\boldsymbol 5}
\def\6{\boldsymbol 6}
\def\7{\boldsymbol 7}
\def\8{\boldsymbol 8}
\def\9{\boldsymbol 9}

\def\E{\mathbb{E}}

\def\P{\mathbb{P}}

\def\R{\mathbb{ R}}

\def\Y{\boldsymbol Y}

\def\AC{\mbox{$\cal A$}}
\def\BC{\mbox{$\cal B$}}
\def\CC{\mbox{$\cal C$}}
\def\DC{\mbox{$\cal D$}}

\def\FC{\mbox{$\cal F$}}
\def\GC{\mbox{$\cal G$}}
\def\HC{\mbox{$\cal H$}}

\def\LC{\mbox{$\cal L$}}

\def\PC{\mbox{$\cal P$}}

\def\SC{\mbox{$\cal S$}}

\def\UC{\mbox{$\cal U$}}
\def\VC{\mbox{$\cal V$}}

\def\|{\Vert}
\def\la{\langle}
\def\ra{\rangle}
\def\1{\mathbbm{1}}

\def\eps{\epsilon}

\def\ti{i'}
\def\tj{j'}
\def\tk{k'}

\def\hm{\widehat \mu}
\def\hg{\widehat \GC}

\usepackage{tikz}
\usetikzlibrary{shapes,arrows}
\usetikzlibrary{calc}

\usepackage{mathtools}
\mathtoolsset{showonlyrefs=true}

\usepackage{subfigure}
\usepackage{caption}

\begin{document}


%
\title{Tensor Kernel Recovery for Discrete Spatio-Temporal Hawkes Processes}
%
%
%
\author{Heejune~Sheen\thanks{H. Sheen is with the Department
of Statistics and Data Science, Yale University, New Haven,
CT, 06511 USA e-mail: heejune.sheen@yale.edu.}, \
        Xiaonan~Zhu\thanks{X. Zhu is with the Department of Operations Research and Financial Engineering, Princeton University, Princeton, NJ 08544, USA e-mail: xz8451@princeton.edu.},
        and~Yao~Xie\thanks{Y. Xie is with H. Milton Stewart School of Industrial and Systems Engineering, Georgia Institute of Technology, Atlanta, GA,30332, USA e-mail: yao.xie@isye.gatech.edu.}
        } 
\maketitle

\begin{abstract}
We introduce a new discrete spatio-temporal Hawkes process model by formulating the general influence of the Hawkes process as a tensor kernel. Based on the low-rank structure assumption of the tensor kernel, we cast the estimation of the tensor kernel as a convex optimization problem using the Fourier transformed nuclear norm. We provide theoretical performance guarantees for our approach and present an algorithm to solve the optimization problem. In particular, our upper bound of squared estimation error has the convergence rate of $O(lnK/\sqrt{K})$, where $K$ is the number of samples in the time horizon.
The efficiency of our estimation is demonstrated with numerical simulations on synthetic data and the analysis of real-world data from Atlanta burglary incidents. 
\end{abstract}


%

\section{Introduction}
%
%
%
%
Hawkes
 processes, a type of self- (and mutual) exciting point processes, have gained substantial attention in machine learning and statistics due to their wide applicability in capturing complex interactions of discrete events over space, time, and possible networks. Such problem arises from many applications such as seismology \cite{ogata1998space}, criminology \cite{zhuang2019semiparametric}, finance \cite{hardiman2013critical,omi2017hawkes}, genomics \cite{reynaud2010adaptive}, and social network \cite{kobayashi2016tideh,zhou2013learning}.  One advantage of Hawkes process modeling is that interactions between the history and a current event can be represented in the structure easily,
as the Hawkes process, in general, has an intensity function consisting of two parts, a baseline intensity, and a triggering effect.

A central problem in Hawkes process modeling is to estimating the triggering effect through the so-called influence functions, which capture how different locations interact with each other. Estimating the triggering effects with Hawkes process models has been conducted in several prior works \cite{INARhawkes,kirchner2018nonparametric,guo2013accelerating,bacry2020sparse,zhou2013learning}. Bacry et al. \cite{bacry2020sparse} and Zhou et al. \cite{zhou2013learning} 
proposed a convex optimization approach with sparse and low-rank assumptions on the interaction adjacency matrix or tensor to estimate an influence in a social network. In particular, they assumed the triggering function as a form of a product between the interaction coefficients and the fixed kernel functions that decay exponentially with continuous time.

Low-dimensional structures are very common in high-dimensional data, such as low-rank matrices and low-rank tensors. A recent motivation for 
studying 
low-rank matrices is due to the matrix completion problem \cite{candes2009exact,candes2010matrix,candes2010power,koltchinskii2011nuclear,cao2015poisson}. One of the popular approaches is convex relaxations with a matrix nuclear norm to estimate a low-rank matrix. There has been much effort in modeling with low-rank tensors by extending the results on low-rank matrices, including \cite{gandy2011tensor,goldfarb2014robust,bahadori2014fast,bengua2017efficient,cai2019nonconvex}. However, unlike matrices, the rank of a tensor is not uniquely defined, and it can have multiple ranks such as the CP rank \cite{carroll1970analysis, harshman1970foundations}, Tucker rank \cite{tucker1966some}, tubal rank, and multi-rank. 
The tubal and multi-rank for the low-rank kernel tensor were proposed by Kilmer $\&$ Martin \cite{kilmer2011factorization} with the algebra for a tensor and the corresponding Fourier-transformed tensor nuclear norm (TNN). 

We are interested in a low-rank structure in this work for the tensor kernel capturing the interaction, which can be viewed as a low-rank approximation to capture the dominant mode of the influence functions. In particular, we consider the tubal and multi-rank for the low-rank kernel tensor.

The main purpose of this paper is to propose a discrete Hawkes process model, which is derived from the spatio-temporal Hawkes process approximation. More precisely, spatio-temporal influence functions for the Hawkes process are first parameterized as a low-rank tensor kernel in our model. Then, an approach to estimate the tensor kernel is presented using maximum likelihood with constrained Fourier transformed nuclear norm on the tensor, which leads to a convex optimization problem. We also prove theoretical performance guarantees for the squared recovery error. It is shown that the squared recovery error of our model converges to 0 at the rate of $O(\ln K/\sqrt{K})$ as the number of samples in time horizon $K$ increases to infinity.
To solve the optimization problem, a computationally efficient algorithm is designed based on the alternating direction method of multipliers (ADMM). 
The computational efficiency of our estimation procedure is illustrated with numerical simulations of synthetic and real data. We emphasize that our approach is different from the previous works \cite{bacry2020sparse,zhou2013learning} since the influence function is considered as a tensor kernel in the discrete-time and discrete location set-up.

The rest of the paper is organized as follows. Section 2 presents our model and the problem setup. Section 3 contains the main theoretical performance upper bound. Section 4 proposes an ADMM-based algorithm to solve the optimization problem. Section 5 contains the numerical study, and finally, Section 6 concludes the paper. The proofs are delegated to the Appendix.

\section{Model}
\subsection{Discrete Hawkes processes}
We first describe continuous spatio-temporal Hawkes processes to motivate our discrete model.
Consider a spatio-temporal point process whose events occur at time $t \in [0,T]$, and the location $(x,y) \in \AC \subset \R^2$. Define a counting process $N : \AC \times [0,\infty) \rightarrow \mathbb{Z}_{>0}$, such that $N(B,C)$ is the number of events in the region $B \in \BC$ and the time window $C \in \CC$, where $\BC$ and $\CC$ are the Borel $\sigma$-algebras of $\AC$ and $[0,\infty)$. Let $\HC_t$ be the $\sigma$-algebra generated by history of the process $N$ up to time $t$. 
The conditional intensity function of a point process is defined as
\begin{align}\label{condinten}
&\lambda(x,y,t) := \lim_{\Delta_x,\Delta_y,\Delta_t \downarrow 0} 
\frac{\E(N([x,x+\Delta_x]\times [y,y+\Delta_y] \times [t,t+\Delta_t] |\HC_t )}{\Delta_x\Delta_y\Delta_t}.
\end{align}
For Hawkes processes, we can define the conditional intensity function with the following form:
\begin{align}\label{Hawkesinten}
&\lambda(x,y,t) = \mu(x,y)
 +\!\int^t_0\!\!\!\iint_{B} \!g(x\!-\!u_1,y\!-\!u_2, t\!-\!u_3) N(d(u_1\!\times\! u_2) \!\times\! du_3),
\end{align}
where $\mu(x,y) \geq 0$ is the base intensity at location $(x,y)$ and $g : \R^2 \times [0.\infty) \rightarrow \R_{\geq 0}$ is the kernel function.

Suppose that the event data lie in bounded region $ [0,n_1\Delta_x] \times [0,n_2\Delta_y]$ and time $[0,K\Delta_t]$ for some $n_1,n_2, K \in \mathbb{Z}_{>0}$. To discretize the process in both space and time, we define  ``bin counts" over  the
discrete space \[\{i,j : 1 \leq i \leq n_1, 1 \leq j \leq n_2 \}\] and time horizon \[\{k : -p+1 \leq k  \leq K \}:\]
\begin{align}\label{bincounts}
\hspace{-0.3cm}  \hspace{-0.1cm}
Z_{ijk} =\hspace{-0.1cm}
    N([(i-1)\Delta_x, i\Delta_x] \times [(j-1)\Delta_y, j\Delta_y] \times [(k-1)\Delta_t, k\Delta_t]) .
\end{align}

Let $\Delta =\Delta_x\Delta_y\Delta_t$. For a given  data preceding discrete time $k$, the expected bin counts can be approximated as follows:
\begin{equation}\label{exapp1}
\begin{aligned}
&\E[Z_{ijk} | \HC_{k-1}]
 \approx \Delta\lambda( (i-1)\Delta_x,(j-1)\Delta_y,(k-1)\Delta_t) \\
&\! \approx  \Delta\mu((i-1)\Delta_x,(j-1)\Delta_y ) \\
&\!+\! \Delta\!\!\!\sum_{\!\tk\!=k\!-\!p}^{k-1}\!\sum_{\ti\!=\!1}^{n_1}\sum_{\tj\!=\!1}^{n_2} g\!\left((i\!-\!\ti)\Delta_x,\!(j\!-\!\tj)\Delta_y,\! (k\!-\!\tk)\Delta_t \right)\!Z_{\!\ti\!\tj\!\tk}\\
&\!:=\! \Delta\!\!\left(\!\mu_{ij}\! +\!\! \!\sum_{\tk=k-p}^{k-1}\sum_{\ti=1}^{n_1}\!\sum_{\tj=1}^{n_2} \GC_{i-\ti\!+n_1,j-\tj\!+n_2,k-\tk}Z_{\ti\!\tj\!\tk\!}\!\right)\!, 
\end{aligned}
\end{equation}
where $\mu \in \R^{n_1\times n_2}$ and $\GC \in \R^{(2n_1-1)\times(2n_2-1)\times p}$ are discretized versions of the base intensity $\mu(x,y)$ and the kernel $g(x,y,t)$, respectively.
For the first approximation above, $\E[Z_{ijk} | \HC_{k-1}]$ is  approximated using  \eqref{condinten} with small $\Delta$. 
For the second approximation, \eqref{Hawkesinten} and \eqref{bincounts} are used to derive the discrete form. 

 It is commonly assumed in literature \cite{bacry2020sparse, zhou2013learning} that $g$ has the following form:
\begin{align} \label{gform}
g(x,y,t) =h(x,y)f(t),
\end{align}
where $f$ is a monotonically decreasing non-negative function, and $f(t)$ goes to zero for large $t$. An example of $f(t)$ is the class of exponential kernels, $f(t)=\alpha e^{-\alpha t}$ . 
For our model, we relax these assumptions so that the kernel function does not need to follow the form \eqref{gform} and $g$ is not necessarily monotonically decreasing non-negative in time $t$.
The history data with memory depth $p$ is instead exploited to approximate the expected current bin counts. Thus, our model can be applied to more general cases.

Now, we propose a discrete spatio-temporal Hawkes process model with the conditional intensity function defined as follows:
\begin{align}
&\lambda_{ijk}(\mu,\GC) := \lambda( (i-1)\Delta_x,(j-1)\Delta_y,(k-1)\Delta_t) 
= \mu_{ij} \!+\!\!\! \sum_{\tk\!=k\!-\!p}^{k-1}\sum_{\ti\!=\!1}^{n_1}\sum_{\tj\!=\!1}^{n_2} \GC_{i-\ti+n_1,j-\tj+n_2,k-\tk}Z_{\ti\tj\tk},
\end{align}
for $1\leq i\leq n_1,1\leq j\leq n_2$, and $1\leq k\leq K$.

Our model is derived from the spatio-temporal Hawkes process.
The structure of our model is different from that of
Kirchner's approximation \cite{INARhawkes} of the temporal Hawkes process with the model INAR($p$). Thus, the proposed model has various advantages over Kirchner's model when analyzing spatio-temporal data.
A more complex setting with higher dimensions (two dimensions in location and one in time) is dealt with in our model, 
and the location space and time-space are simultaneously discretized with tensor $\GC$. 
Moreover, our interpretation of the discretized version of the kernel function enables the presence of space-time interactions. With constraints imposed on the rank of the tensor and entry-wise bounds on the estimators, a better estimation of the Hawkes process can be obtained when its coefficients are low-rank. Consequently, a convex optimization problem is constructed based on the likelihood function and our corresponding regularization,
 while \cite{INARhawkes} employed the projection method on the approximated time series model INAR($p$). 

To estimate the base intensity matrix $\mu$ and the underlying tensor $\GC$, the followings are assumed: First, we assume that each entry of $\mu$ and $\GC$ has the upper and the lower bound, i.e., there exist non-negative constants $a_1,b_1,a_2,b_2$ such that $a_1\leq\mu_{ij}\leq b_1$, $a_2\leq\GC_{ijk}\leq b_2$ and $a_1+a_2>0$. This assumption ensures that our problem is well-posed. 

\begin{figure}[t]
\centering
\includegraphics[width=.9\columnwidth]{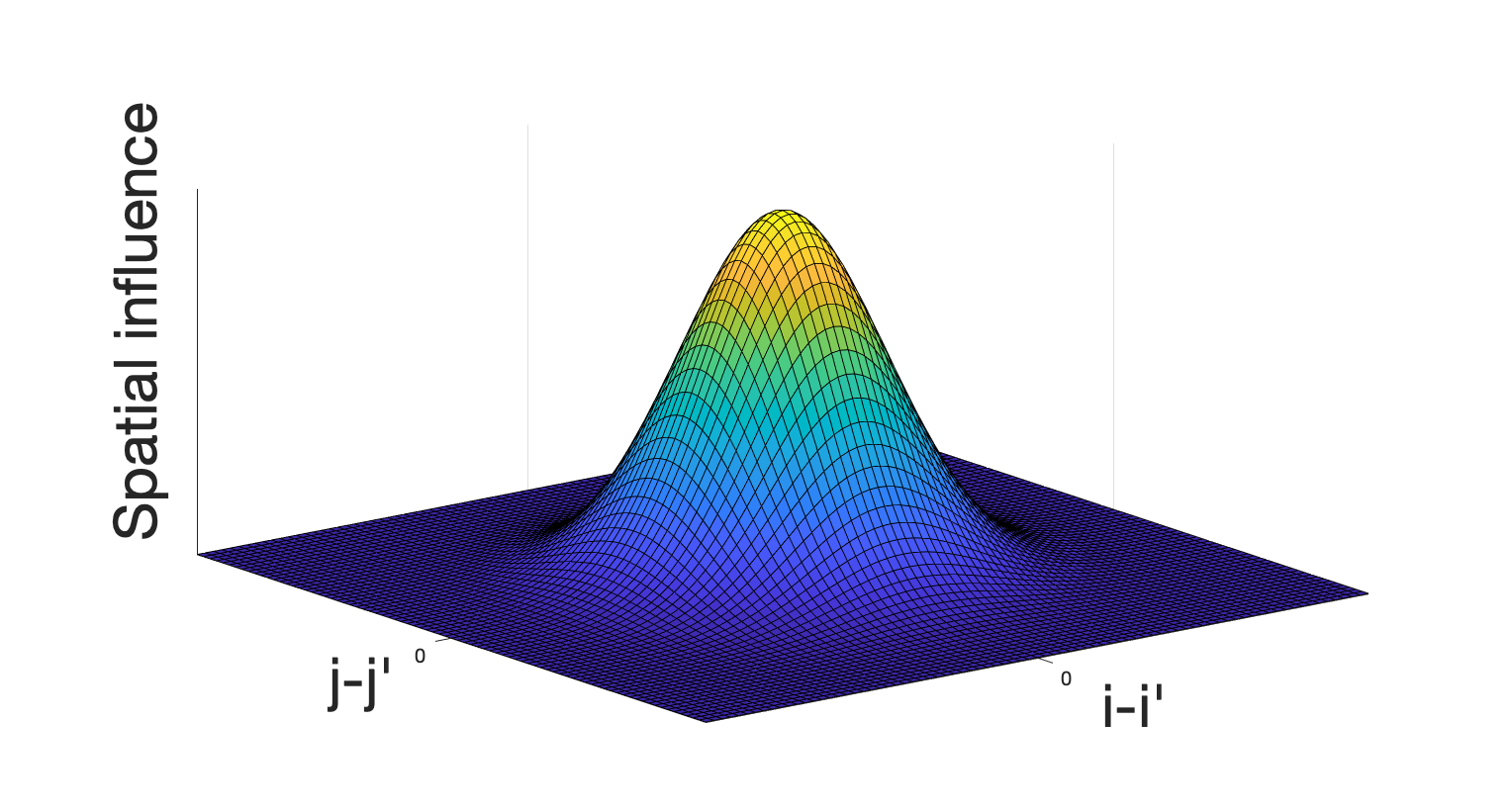}
\caption{For any given $(i,j)$ grid, it displays 
the case where the spatial influence from $(i,j)$ to $(i',j')$ follows a standard gaussian function.}
\label{Fig_lowrankex}
\end{figure}

Second, we also assume that 
the tensor $\GC$  has a low multi-rank $(r_1, \ldots , r_p)$, where $r_k = {\rm rank}(\widetilde\GC^{(k)})$ and $\GC^{(k)}$ is the $k$th frontal slice of transformed tensor $\widetilde\GC$ (i.e., $\GC^{(k)}:=\GC(:,:,k)$ using MATLAB notation).  The transformed tensor $\widetilde \GC$ is obtained by applying the Discrete Fourier Transform (DFT) to the mode-3 fibers of $\GC$ (Lemma \eqref{FourierDonain}). In other words,   a small sum of multi-rank $\gamma := \sum_{i=1}^{p} r_k$ is assumed.
This assumption 
is based on the high correlations that exist within locations and time. 
For instance, for any given grid $(i,j)$, suppose that the spatial influence from $(i,j)$ to $(i',j')$ is proportional to a standard Gaussian function (i.e., $ \propto e^{-((i-i')^2+(j-j')^2)/2}$) as illustrated in Figure \ref{Fig_lowrankex}, and the temporal influence follows a decreasing function in time. 
 Then, the tensor $\GC \in \R^{(2n_1-1)\times(2n_2-1)\times p}$ in our model has a low multi-rank at most $(1, \ldots,1)$ and a small sum of  the multi-rank less or equal to $p$.

Our problem considers a transformed multi-rank of a kernel tensor and a TNN over other tensor ranks and norms.
The rank of a tensor can be defined in several ways, for instance, the  CP rank,  the Tucker rank, 
and the multi-rank. 
Depending on the tensor rank,  
the corresponding tensor nuclear norms should be utilized.
It is known that the computation of the CP rank 
 is NP-hard \cite{hillar2013most} and its relaxation is intractable in general. For this reason, the tractable Tucker rank and its relaxation are usually used.
 One of the popular relaxations is the sum of the matrix nuclear norms of matrices obtained by unfolding a tensor \cite{liu2012tensor}. 
It is, however, not the tightest convex relaxation of the Tucker rank \cite{romera2013new} and the matrix norm may be inefficient when the unfolding matrices have a significant difference in the number of rows and columns.  
We use TNN since it is the convex envelope of the multi-rank and can successfully interpret the low-rank structure of a kernel tensor.

\subsection{Low-rank tensor regularization}

In this section, we review the tensor nuclear norm (TNN), which is used to guarantee the low-rankness of $\GC$.
We begin by introducing some notation. For matrix $X$, let $\|X\|$ be the matrix spectral norm, $\|X\|_*$  the matrix nuclear norm, and  $\|X\|_F=(\sum_{i,j}X_{ij}^2)^{1/2}$  the Frobenius norm. 
For 3-way tensor $\GC\in \mathbb{R}^{N_1\times N_2\times N_3}$,  MATLAB notation is used to denote $k$-th horizontal, lateral, and frontal slice by $\GC(k,:,:)$, $\GC(:,k,:)$, and $\GC(:,:,k)$ respectively. Specifically, the $k$-th frontal slice of $\GC$ is denoted by $\GC^{(k)}:=\GC(:,:,k)$ for $k=1,\ldots,N_3$. The $k$-th mode-3 fiber of a 3-way tensor is defined by holding the first two indices fixed and varying the third, and denoted by $\GC(k,k,:)$. 
The norm of tensor is defined as $\|\GC\|_1 = \sum_{i,j,k}|\GC_{ijk}|$, $\|\GC\|_F = (\sum_{i,j,k}\GC_{ijk}^2)^{1/2}$. The tensor spectral norm $\|\GC\|_{spec}$ is defined later in Definition \ref{def_TNN}.

We introduce the following operators for the tensor algebra: the block circulation, the block diagonalization, and the fold and unfold command of tensor $\GC$. 
\begin{align}
&\textsf{bcirc}(\GC) \!=\! 
\left(\!\!\begin{array}{ccccc} 
\GC^{(1)} &\GC^{(N_3)} &\GC^{(N_3-1)} &\cdots &\GC^{(2)}  \\ 
\GC^{(2)} &\GC^{(1)} &\GC^{(N_3)} &\cdots &\GC^{(3)} \\ 
\vdots &\ddots &\ddots &\ddots &\vdots \\ 
\GC^{(N_3)} &\GC^{(N_3-1)} &\GC^{(N_3-2)} &\cdots &\GC^{(1)}
\end{array}\!\!\right)\!,
\end{align}
\vspace{-3mm}
\begin{align}
&\textsf{blockdiag}(\GC)=\left(\begin{array}{cccc} 
\GC^{(1)} & & & \\ 
&\GC^{(2)} & & \\ 
& & \ddots & \\ 
& & &\GC^{(N_3)}
\end{array}\right),
\end{align}
\vspace{-3mm}
\begin{align}
\textsf{unfold}(\GC) \!=\! 	
\left(\!\begin{array}{c} 
\GC^{(1)} \\ \GC^{(2)} \\ \vdots \\ \GC^{(N_3)}
\end{array}\!\right),
\ \text{ and }\ \ 
\textsf{fold}(\textsf{unfold}(\GC)) \!=\! \GC.
\end{align}

For two tensors $\GC_1\!\in\! \R^{N_1\times N_2 \times N_3}$ and $\GC_2 \!\in\! \R^{N_2\times N_4 \times N_3}$, the {\it t-product} is defined as
\begin{align}
\GC_1 * \GC_2= \textsf{fold}(\textsf{bcirc}(\GC_1)\textsf{unfold}(\GC_2))\in\mathbb{R}^{N_1\times N_4\times N_3}.
\end{align}

Note that Kilmer and Martin \cite{kilmer2011factorization} proposed a singular value decomposition (SVD) method for three-way tensors, and based on the tensor SVD, TNN is proposed by Semerci et al. \cite{semerci2014tensor}.
We first review some background materials on the tensor SVD to introduce TNN.
See \cite{kilmer2011factorization} for more information.
For a tensor $\GC \in \R^{N_1\times N_2 \times N_3}$, 
the block diagonalization property of block circulant matrices is described in the following lemma.
\begin{lemma} {\upshape{\cite{kilmer2011factorization}}}\label{FourierDonain}
For $\GC\! \in\! \R^{N_1\times N_2 \times N_3}$,  $\textsf{bcirc}(\GC) \!\in\! \R^{N_3N_1\times N_3N_2}$, we have 
\begin{align}
(F_{N_3} \otimes I_{N_1} ) \textsf{bcirc}(\GC) &(F_{N_3}^* \otimes I_{N_2})=\textsf{blockdiag}(\widetilde \GC),
\end{align}
where $\otimes$ is the Kronecker product, $I_{N_1}$ and $I_{N_2}$ are identity matrices in $\R^{N_1 \times N_1}$ and $\R^{N_2 \times N_2}$, respectively, $F_{N_3} \in \R^{N_3 \times N_3}$ is the normalized DFT matrix, which is unitary, and $F^*$ denotes its conjugate transpose.
The matrix $\textsf{blockdiag}(\widetilde \GC)$ is the transformation of $\textsf{bcirc}(\GC)$ into the Fourier domain, and the tensor $\widetilde \GC$ is obtained by performing the DFT to the mode-3 fibers of $\GC$ as mentioned earlier.
\end{lemma}

Based on the matrix SVD, we have 
\begin{equation}\label{Fourier_svd}
\textsf{blockdiag}(\!\widetilde \GC)\!\!=\!\!\textsf{blockdiag}(\!\widetilde \UC)\textsf{blockdiag}(\!\widetilde \SC)\textsf{blockdiag}(\!\widetilde \VC),
\end{equation}
 in the Fourier domain,
where $\widetilde \GC^{(k)} = \widetilde \UC^{(k)} \widetilde \SC^{(k)} (\widetilde \VC^{(k)})^\top$ is the SVD.
The equivalent decomposition of three-way tensors to \eqref{Fourier_svd} is characterized as a tensor-SVD \cite{zhang2016exact}.
The tensor-SVD for three-way tensors is described as follows.
\begin{theorem}\label{svd}
\upshape{\cite{kilmer2011factorization}}
\itshape{Any tensor $\GC \in \R^{N_1\times N_2 \times N_3}$ can be factored as
\setlength\belowdisplayskip{0.8pt}
\begin{align}
\GC &= \UC * \SC * \VC^\top 
= \sum_{i=1}^{N_1\wedge N_2} \UC(:,i,:) * \SC(i,i,:) * \VC(:,i,:)^\top,
\end{align}
}
\end{theorem}

Combining Theorem \ref{svd} and equation \eqref{Fourier_svd}, TNN can be defined as follows.
 
\begin{definition}
[\upshape{Theorem 2.4.1 in \cite{zhang2014novel}}]\label{def_TNN}
The tensor nuclear norm (TNN) of $\GC$ is defined as the sum of the singular values of all the frontal slices of $\widetilde \GC$:
\begin{align}
\| \GC \|_{\rm TNN} = \sum_{i=1}^{N_1 \wedge N_2}\sum_{j=1}^{N_3} \widetilde \SC_{i,i,j}.
\end{align}
Note that the dual norm of the tensor nuclear norm is the tensor spectral norm $\| \GC \|_{\rm spec} := \|\textsf{bcirc}(\GC) \|$.
\end{definition}

\subsection{Problem formulation}
We use the notation:
\begin{align}
Z^{t} &:= \{Z_{ijk}, 1 \leq i \leq n_1, 1 \leq j \leq n_2, -p+1\leq k \leq t \},  \\
Z^{t}_{q} &:= \{Z_{ijk}, 1 \leq i \leq n_1, 1 \leq j \leq n_2, q\leq k \leq t \}.
\end{align}
Assume that the discrete data follow the Poisson distribution:
\begin{align} \label{Data}
 Z_{ijk} | \HC_{k-1} \sim {\rm Poisson}\left(\Delta\lambda_{ijk}\right).
 \end{align}
Note that for fixed $k$, $Z_{ijk}$ are conditionally independent for all $i,j$ given $\HC_{k-1}$, and $\lambda_{ijk}$ depends only on the history of data before $k$, not on the data at time $k$. We also mention that our analysis can be applied to the alternative, the Bernoulli distribution assumption with slight modifications.
Our goal is to estimate the true parameters $\mu$ and $\GC$ of the discrete Hawkes process model. 
By our assumptions on the low-rank tensor $\GC$, we have
\begin{align}
\|\GC\|_{\rm TNN} &= \|\textsf{bcric}(\GC)\|_* = \|\textsf{blockdiag}(\widetilde\GC)\|_* \\
&\leq\! \sqrt{\gamma}\|\textsf{blockdiag}(\widetilde\GC) \|_F =\sqrt{\gamma}\|\textsf{blockdiag}(\GC) \|_F \\
&\leq b_2\sqrt{\gamma(2n_1-1)(2n_2-1)p}.
\end{align}
Accordingly, the candidate set $\DC$ for the true value $(\mu,\GC)$ is defined as:
\begin{align}
\DC := \{ (\mu ,\GC) &| \ a_1 \leq \mu_{ij} \leq b_1, \ a_2 \leq \GC_{ijk} \leq b_2, \\
& \|\GC\|_{\rm TNN} \leq b_2\sqrt{\gamma(2n_1-1)(2n_2-1)p} \ \}.
\end{align}

We consider a formulation by maximizing the log-likelihood function of the optimization variable $\mu$ and $\GC$ given the observations $Z^K$. The negative log-likelihood function is given by
\begin{equation} \label{M-est}
F\!({\mu}, {\mathcal{G}}) \!\!:=\!\!  \sum_{k=1}^{K}\!\sum_{i=1}^{n_1}\!\sum_{j=1}^{n_2} ( \!\Delta\!\lambda_{ijk\!}(  {\mu}, {\mathcal{G}}) -\! Z_{ijk\!}\ln(\!\Delta\!\lambda_{ijk\!}({\mu}, {\mathcal{G}}))).
\end{equation}
Therefore, the estimators $(\widehat{\mu}, \widehat{\mathcal{G}})$ can be obtained by solving the following convex optimization problem:
\begin{equation}\label{P1}
\begin{aligned}
(\widehat{\mu}, \widehat{\mathcal{G}})
		=\argmin_{(\mu ,\GC)\in \DC} F(\mu,\GC).
\end{aligned}
\end{equation}

\begin{rema}
	Note that the convex optimization problem \eqref{P1} constrained in a candidate set $\DC$ can also be formulated as a regularized maximum likelihood function problem. Indeed, there exists a constant $\tau \in \mathbb{R}$ such that problem \eqref{P1} equals 
	\[
		(\widehat{\mu}, \widehat{\mathcal{G}})
		=\argmin\left\{ F(\mu,\GC)+\tau\|\GC\|_{\rm TNN}\right\}
	\]
	with the natural constraint upon the entries:
	\[
	 a_1 \leq \mu_{ij} \leq b_1, \ a_2 \leq \GC_{ijk} \leq b_2, \ \forall\, i,j,k.
	\]
	It follows from the duality theory in optimization \cite{boyd_vandenberghe_2004}. We use the regularized form to derive the algorithm in Section \ref{SecAlgrithm}.
\end{rema}

\section{Theoretical guarantee}
We present an upper bound for the sum of squared errors of the two estimators, which is defined by
\begin{align}
R[(\mu,\GC) || (\widehat{\mu},\widehat{\GC}) ] :=  \|\mu-\widehat{\mu}\|_F^2 + \| \GC-\widehat{\GC} \|_F^2,
\end{align}
where $\widehat{\mu}$ and $\widehat{\GC}$ are the optimal solutions to \eqref{P1}.

To state our theoretical guarantee, we define the condition number as in \cite{juditsky2020convex}.
\begin{definition}\label{Def2}
Given $X \in \R^{d\times d}, X \succeq 0$ and $p \in [1,\infty]$, the condition number is defined by
\[
\delta_{p}[X] := \max \{ \delta \geq 0: g^\top X g \geq \delta\|g\|_{p}^2, \ \forall g \in \R^d \},
\]
where  $X \succeq 0$ ($X \succ 0$)  denotes that $X$ is a positive semidefinite matrix (a positive definite matrix, respectively).
Note that $\delta_{p}[X] >0$ when $X \succ 0$. 
\end{definition}

Now, we present our main theorem.

\begin{theorem} {\upshape(Estimation error driven by data)} \label{Thm1}
Assume that $(\mu,\GC) \in \DC$ and $(\widehat{\mu},\widehat{\GC})$ are the optimal solution to \eqref{P1}. 
Let 
\begin{align*}
 \underbar{$J$} &= a_1+a_2\min_{k}\{||Z^{k-1}_{k-p}||_1 \},\\
\bar J & = b_1+b_2\sqrt{\gamma(2n_1-1)(2n_2-1)p}\max_{1 \leq k \leq K}\{ \| Z^{k-1}_{k-p} \|_{\rm spec}\},
\end{align*}
and let $A[\,\cdot\,]:\, \mathbb{R}^{n_1\times n_2\times (K+p)}\rightarrow \mathbb{R}^{d\times d}$ be a mapping defined in Appendix A. 
Then, for every $Z^K$, $\alpha_1,\alpha_2 \in (0,1)$, it holds that 
\setlength\belowdisplayskip{1pt}
\begin{align}
 &  R[(\mu,\GC) || (\hm,\hg) ] \label{MUB} \\
&\!\leq\! \frac{16\sqrt{2} n_1n_2 \bar J^2  \ln ( \bar J / \underbar{$J$} )} {\sqrt{K}(1\!-\!e^{-2 \bar J})\Delta\delta_{2}[A[Z^K]]} \sqrt{ \ln\frac{n_1n_2}{\alpha_2} }\\
&\qquad \qquad \cdot \max \left \{ 2\sqrt{\Delta \bar J\ln\frac{n_1n_2K}{\alpha_1} } , \  4 \ln\frac{n_1n_2K}{\alpha_1}    \right \}\end{align}
with probability at least $1-2\alpha_1 -2\alpha_2$.
\end{theorem}

\begin{rema}\label{nonrandomUP}
If $\delta_{2}[A[Z^K]]> 0 $, for large $n_1,n_2,p$ and $K$, there exists a constant $C>0$ such that 
the following bound holds with high probability. 
\begin{align*}
   R[(\mu,\GC) || (\hm,\hg) ] 
\leq C\frac{ n_1n_2 \bar J^2  \ln ( \bar J )\sqrt{ \ln (n_1n_2) }\cdot  \ln (n_1n_2K)  }{\sqrt{K}}.   
\end{align*}

\end{rema}

\begin{rema}
From Remark \ref{nonrandomUP}, we observe that, for given data $Z^K$, the upper bound can be regarded as an increasing function of $\bar J$. More precisely, the estimation error for the upper bound increases with the upper bound on the tensor nuclear norm $b_2[\gamma(2n_1-1)(2n_2-1)p]^{1/2}$. It implies that the upper bound of the estimation error will be small if we have a small sum of multi-rank $\gamma$. It is a characteristic that we can expect from the low-rank tensor recovery. 
\end{rema}

\begin{figure}[t] 
\centering
\includegraphics[width=.8\columnwidth]{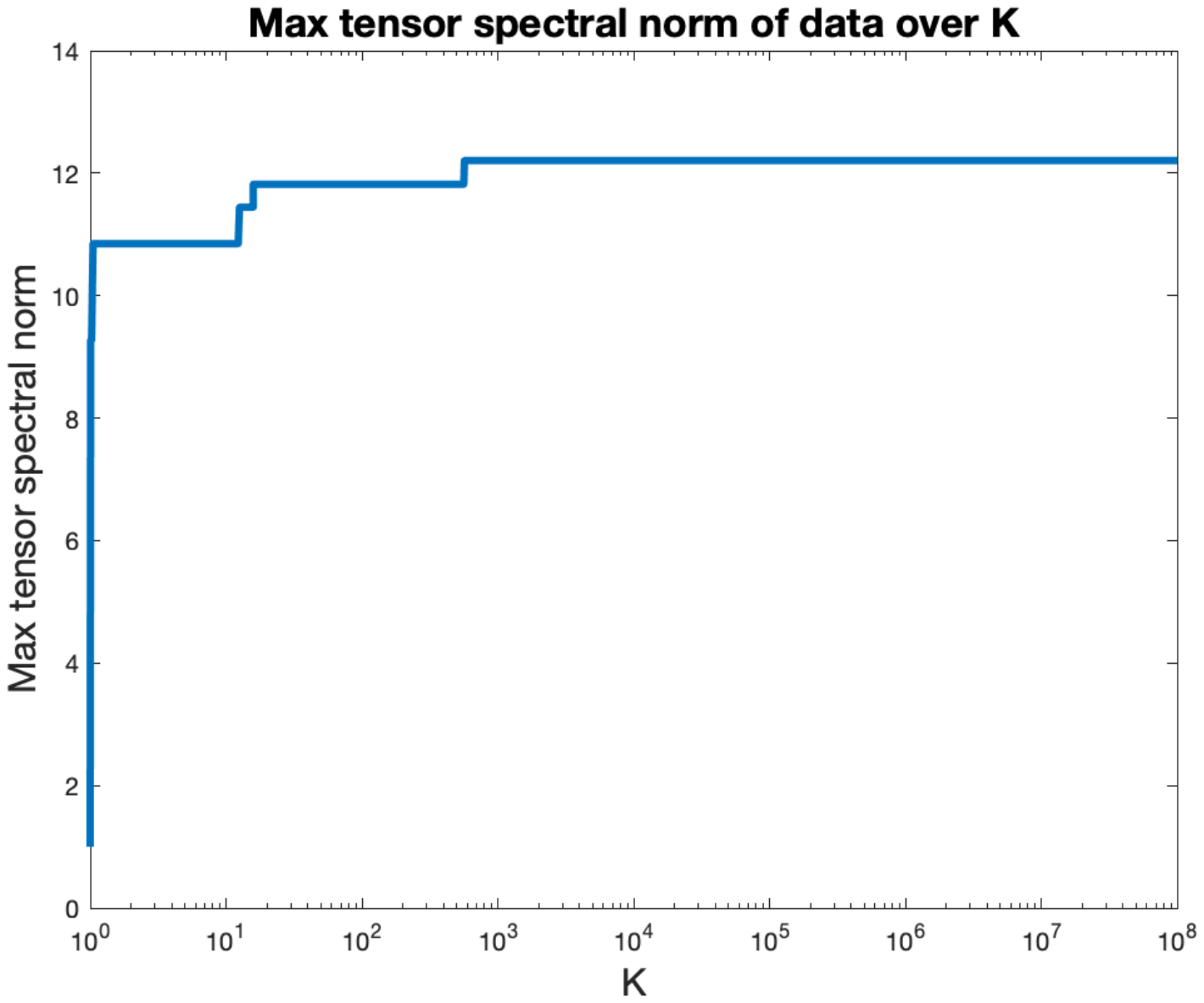}
\caption{The simulation of $\max_{k} \| Z^{k-1}_{k-p} \|_{spec}$}
\label{specnorm}
\end{figure}

\begin{rema} \label{convergeR}
We observe that by  fixing, $n_1$, $n_2$, $\Delta$, $a_1$, $a_2$, $b_1$, $b_2$, and $\gamma,$ the upper bound \eqref{MUB} tends to $0$ as $K \rightarrow \infty$ at the rate 
of  $O(\ln K/\sqrt{K})$. 
We experimentally show that $\max_{1 \leq k \leq K}\{ \| Z^{k-1}_{k-p} \|_{\rm spec}\}$ is bounded above by O(lnK)  in Figure \ref{specnorm}.
\end{rema}

\noindent

\noindent
The proof for Theorem \ref{Thm1} is presented in Appendix A. In the proof,  
the Kullback-Leibler (KL) divergence and Hellinger distance are defined between two Poisson distributions.
 For any two Poisson mean $p$ and $q$,  the KL -divergence is defined as
\begin{align}
D(p||q ) := p \ln(p/q) -( p-q ),
\end{align}
and the Hellinger distance as
\begin{align}
H^2(p||q) := 2-2\exp(-\frac{1}{2}(\sqrt{p} -\sqrt{q})^2).
\end{align}
Then, a lower bound is derived for 
$\sum_{k=1}^{K}\sum_{i=1}^{n_1}\sum_{j=1}^{n_2}  D(\lambda_{ijk}(\mu,\GC) || \lambda_{ijk}(\hm,\hg))$ with the Hellinger distance and Lemma 8 in \cite{cao2016poisson}. Furthermore, we establish an upper bound on the sum of the KL divergence using the Azuma Hoeffding's inequality. 
We then obtain the upper bound for the estimation error by combining the lower and upper bound. 

Corollary \ref{Corr} immediately follows from Theorem \ref{Thm1}. 
 In particular, it demonstrates the data-driven upper bound for the sum of KL divergence between the estimated and the true intensity functions. 
\begin{coro}\label{Corr}
Assume that $(\mu,\GC) \in \DC$ and $(\hm,\hg)$ are the optimal solution to \eqref{P1}. With the notation defined in Theorem \ref{Thm1},
for every $Z^K$, $\alpha_1,\alpha_2 \in (0,1)$, it holds that 
\setlength\belowdisplayskip{1pt}
\begin{align} 
 &  \sum_{k=1}^{K}\sum_{i=1}^{n_1}\sum_{j=1}^{n_2}  \frac{D(\lambda_{ijk}(\mu,\GC) || \lambda_{ijk}(\hm,\hg))}{n_1n_2K} 
\leq  \sqrt{\frac{2}{K}} \ln \frac{ \bar J}{\underbar{$J$} } \ln\frac{n_1n_2}{\alpha_2} 
 \max \left \{ 2\sqrt{\Delta \bar J\ln\frac{n_1n_2K}{\alpha_1} } , \  4 \ln\frac{n_1n_2K}{\alpha_1}    \right \}
\end{align}
with probability at least $1-2\alpha_1 -2\alpha_2 $.
\end{coro}

\section{Algorithm}\label{SecAlgrithm}
For the proposed convex optimization problem \eqref{P1}, 
we apply ADMM and the majorization-minimization (MM) algorithms.
Based on the \texttt{ADM4} algorithm proposed by \cite{zhou2013learning}, 
we design our algorithm for problem \eqref{P1}. 
To start with, the constraint sets for $\mu$ and $\GC$ are separated to the following two closed convex sets:
\begin{align}
&\Gamma_1 \!:=\! \{ \mu \ \!| \ a_1 \leq \mu_{ij} \leq b_1, \ \forall (i,j) \in \llbracket n_1 \rrbracket \!\times\! \llbracket n_2 \rrbracket \}, \\
&\Gamma_2 \!:=\! \{ \GC \ \!| \ a_2 \leq \GC_{ijk} \leq b_2, \ \\
& \qquad \qquad \qquad \forall (i,j,k) \in \llbracket 2n_1-1 \rrbracket \!\times\! \llbracket 2n_2-1 \rrbracket \!\times\! \llbracket p \rrbracket  \},
\end{align}
where $\llbracket n \rrbracket := \{1,2,...,n\}$.
Then, problem \eqref{P1} can be written as
\begin{align} \label{AP1}
\min \ \ \ & \ F(\mu,\GC) + \tau\|\GC\|_{\rm TNN} \\
\mbox{subject to} \ \ \  & \ \mu \in \Gamma_1, \ \GC \in \Gamma_2.
\end{align}
ADMM is employed to convert the above optimization problem to several sub-problems that are easier to solve. More specifically, the problem is separated into the first term of the objective function, the regularization term, and the constraints. To that end, three auxiliary variables, $m,G$ and $R$ are introduced, and \eqref{AP1} can be equivalently expressed as
\begin{align} \label{AP2}
\min \ \ \ & \ F(\mu,\GC) + \tau\|R\|_{\rm TNN} \\
\mbox{subject to} \ \ \  & \ m \in \Gamma_1, \ G \in \Gamma_2, \\ 
&\ \mu = m, \ \GC= G, \ \GC = R. 
\end{align}
We then define the following augmented Lagrangian function of \eqref{AP2}:
\begin{align}
 &\LC_\rho(\GC, \mu, R, G, m, Y_1,Y_2,Y_3) \\
 &:= F(\mu, \GC) + \tau \|R\|_{\rm TNN}+ \psi_{\Gamma_1}(m) + \psi_{\Gamma_2}(G) \\
 &\quad+ \rho\la Y_1,\GC-R\ra +\rho\la Y_2,\GC-G\ra  + \rho\la Y_3,\mu-m \ra \\
 &\quad+\frac{\rho}{2}\|\GC-R\|_F^2 +\frac{\rho}{2}\|\GC-G\|_F^2   +\frac{\rho}{2}\|\mu-m\|_F^2, 
\end{align}
where $Y_1,Y_2$, and $Y_3$ are the dual variables associated with the constraints $\GC = R$, $\GC= G$, and $\mu = m$, respectively. The constant $\rho$ is a penalty parameter, 
and the functions $\psi_{\Gamma_1}(m)$ and $\psi_{\Gamma_2}(G)$ are defined as  
\begin{align}
\psi_{\Gamma_1}(m) &:=
  \begin{cases}
    0 &    \ \text{if} \ m \in \Gamma_1,\\
    +\infty &\text{otherwise}, 
  \end{cases} \\
  \psi_{\Gamma_2}(G) &:=
  \begin{cases}
    0 &    \ \text{if} \ G \in \Gamma_2,\\
    +\infty &\text{otherwise}. 
  \end{cases}
\end{align}
Notice that two blocks of variables $\left(\GC^{t+1},\mu^{t+1}\right)$ and $(R^{t+1},m^{t+1},G^{t+1})$ are separable in the augmented Lagrangian function. Thus, ADMM can be applied as the following iterations.
\begin{flalign}
\left(\GC^{t+1},\mu^{t+1}\right) &= \argmin_{\mu, \GC} \LC_{\!\rho}(\GC,\! \mu,\! R^t,\! G^t,\! m^t,\! Y_1^t,\! Y_2^t,\! Y_3^t), \label{step1} \\
(R^{t+1},m^{t+1},G^{t+1})  
 & = \argmin_{R,m,G} \LC_{\!\rho}(\GC^{t+1},\! \mu^{t+1},\! R,\! G,\! m,\! Y_1^t,\! Y_2^t,\! Y_3^t), \label{step2} \\
 & Y_1^{t+1}  = Y_1^t+ (\GC^{t+1}-R^{t+1}), \\
 & Y_2^{t+1}  = Y_2^t+ (\GC^{t+1}-G^{t+1}), \\
 &Y_3^{t+1} = Y_3^t+ (\mu^{t+1}-m^{t+1}). &&
\end{flalign}
It has been shown in \cite{fazel2013hankel} that the above application of ADMM on the two-block convex minimization problem converges.

It remains to solve \eqref{step1} and \eqref{step2} respectively.
We start with deriving the updating step for $\GC$ and $\mu$ as follows.
Note that for \eqref{step1}, the following optimization problem is considered:
{\setlength\abovedisplayskip{1pt}
\setlength\belowdisplayskip{1pt}
\begin{align} 
\min_{\mu ,\GC}&\ g(\mu,\GC) :=  \sum_{\tk=1}^{K}\sum_{\ti=1}^{n_1}\sum_{\tj=1}^{n_2} \left[ \Delta\lambda_{\ti\tj\tk}(\mu,\GC)\right. \\
& \left.- Z_{\ti\tj\tk}\ln(\Delta\lambda_{\ti\tj\tk}(\mu,\GC))\right] + \frac{\rho}{2}\| \GC - R^t + Y_1^t \|_F^2 \\
&+ \frac{\rho}{2}\| \GC - G^t + Y_2^t \|_F^2 + \frac{\rho}{2}\| \mu - m^t + Y_3^t \|_F^2. \label{step1_1}
\end{align}
}
Since no closed-form solutions exist, we apply the MM algorithm as in \cite{zhou2013learning}.
For any $\mu, \GC$, let $Q(\mu, \GC ;  \mu^{(q)}, \GC^{(q)})$ be a convex function such that
\begin{align}
g(\mu,\GC) &\leq Q(\mu, \GC ;  \mu^{(q)}, \GC^{(q)}) \label{Q1}\\ 
g( \mu^{(q)}, \GC^{(q)}) &= Q( \mu^{(q)}, \GC^{(q)} ;  \mu^{(q)}, \GC^{(q)}), \label{Q2}
\end{align}
where $ \mu^{(q)}$ and $\GC^{(q)}$ are estimates of $\mu$ and $\GC$. Then, we can obtain the optimal solutions 
to convex problem \eqref{step1_1} by using the iterative procedure:
\begin{align}
\left(\mu^{(q+1)}, \GC^{(q+1)}\right)= \argmin_{\mu, \GC} Q(\mu, \GC ;  \mu^{(q)}, \GC^{(q)}).
\end{align}
Let 
\begin{align}
\Omega = \{ k \ | \ Z_{ijk} \neq 0 \ \text{for some} \ i \ \text{and} \  j \}
\end{align}
and
\[
l(k) = \{ (i,j) \ | \ Z_{ijk} \neq 0 \}.
\]
Define $Q(\mu, \GC ;  \mu^{(q)}, \GC^{(q)})$ that satisfies \eqref{Q1} and \eqref{Q2}  as follows:
\begin{align}
& Q(\GC,\mu ; \GC^{(q)}, \mu^{(q)}) \\
&= \! -\!\sum_{\tk \!\in \Omega}\sum_{(\ti\!,\tj\!)  \in l(\tk\!)} \!\left[ Z_{\ti\! \tj\! \tk\!}\ln \Delta \!+\! Z_{\ti\! \tj\! \tk}\!\left(\!p_{\ti\tj\tk}\ln\frac{\mu_{\ti \tj}}{p_{\ti\tj\tk}} \right.\right.\\
& \left.\left.\!+\!\! \sum_{k=\tk\!\!-\!p}^{\tk-1}\sum_{(i,j) \in l(k)} \!\!\!p_{ijk,\ti\! \tj\! \tk}\ln\frac{\GC_{\ti\!-i+n_1,\tj\!-j+n_2,\tk\!-k }Z_{ijk}}{p_{ijk,\ti\! \tj\! \tk}} \!\right) \!\right] \\
&+\! \!\sum_{\tk\!=1}^{K}\sum_{\ti\!=1}^{n_1}\sum_{\tj\!=1}^{n_2} \!\Delta\!\!\left(\!\sum_{k=\tk\!\!-\!p}^{\tk-1}\sum_{(i,j)\in l(k)} \!\!\!\!\GC_{\ti\!-i+n_1,\tj\!-j+n_2,\tk\!-k}Z_{ijk\!}\!\right) \\
&+ \frac{\rho}{2}\| \GC - R^t + Y_1^t \|_F^2 + \frac{\rho}{2}\| \GC - G^t + Y_2^t \|_F^2 \\
&+ \frac{\rho}{2}\| \mu - m^t + Y_3^t \|_F^2+n_3\Delta \sum_{\ti=1}^{n_1}\sum_{\tj=1}^{n_2} \mu_{\ti \tj},
\end{align}
where
\begin{align}
p_{\ti\!\tj\!\tk} \!=\! \frac{\mu_{\ti\! \tj\!}^{(q)}}{\lambda_{\ti\! \tj\! \tk}^{(q)}}, \mbox{ and \ } 
p_{ijk\!,\ti\! \tj\! \tk} \!=\! \frac{\GC_{\ti\!-i+n_1\!,\tj\!-j+n_2\!,\tk\!-k}^{(q)}Z_{ijk}}{\lambda_{\ti\! \tj\! \tk}^{(q)}}.
\end{align}
Let $a= \ti-i+n_1$, $b=\tj-j+n_2$, $c=\tk-k$, and $l'\!(k) \!=\! l(k)\cap \{\ti\! -i+n_1\!=\!a, \tj\! -j+n_2\!=\!b, \tk\!-k\!=\!c\}$. By taking derivative, a closed form solution to $Q(\mu, \GC ;  \mu^{(q)}, \GC^{(q)})$ is 
\begin{align}
\mu_{\ti \tj}^{(q+1)} &=  \frac{-B +\sqrt{B^2-4\rho C}}{2\rho},\label{MM1} \\
\GC_{abc}^{(q+1)} &=  \frac{-U +\sqrt{U^2-8\rho V}}{4\rho}, \label{MM2}
\end{align}
where 
\begin{align}
B &= n_3\Delta +\rho\big[-(\mu^t)_{\ti \tj} +(Y_3^t)_{\ti \tj}\big],\\
C &= -\sum_{\tk \in \Omega}\sum_{(\ti,\tj) \in l(\tk)}Z_{\ti \tj \tk}p_{\ti\tj\tk}, \\
U &= \sum_{\tk=1}^{K}\sum_{\ti=1}^{n_1}\sum_{\tj=1}^{n_2}\left(\sum_{k=\tk-p}^{\tk-1}\sum_{(i,j)\in l'(k)} \Delta Z_{ijk}\right) \\
&\quad+\rho\big[-(R^t)_{abc} +(Y_1^t)_{abc} -(G^t)_{abc} +(Y_2^t)_{abc}\big], \\
V &= -\sum_{\tk \in \Omega}\sum_{(\ti,\tj) \in l(\tk)} \sum_{k=\tk-p}^{\tk-1}\sum_{(i,j)\in l'(k)} Z_{\ti \tj \tk} p_{ijk,\ti \tj \tk}.
\end{align}

Recall that $R, m$, and $G$ in the objective function of the problem \eqref{step2}  are separable. Hence they can be calculated one by one. We start with updating step $R$.
The optimal solution to \eqref{step2} regarding $R$ is given by
\begin{align} \label{Req}
R^{t+1} &= \argmin_{R} \tau\| R \|_{\rm TNN} +\frac{\rho}{2} \| -R_1 + Y_1^t + \GC^{t+1} \|_F^2 \\
&= {\rm Prox}_{(\tau/\rho)\| \cdot \|_{\rm TNN}}\left(Y_1^t + \GC^{t+1} \right) \\
&= \UC * \SC_{\rho/\tau} * \VC^\top,
\end{align}
where  $\UC * \SC * \VC^\top$ is a tensor singular value decomposition of $Y_1^t + \GC^{t+1}$, $\SC_{\rho/\tau} =\texttt{IFT}(\widehat \SC_{\rho/\tau},[\ ],3)$ for the third frontal slices, and $\widehat \SC_{\rho/\tau} :=\max \{\widehat S -\rho/\tau ,0\}$. Here, the operator $\texttt{IFT}$ corresponds to an inverse Fourier transform.

For updating $G$, an optimal solution to problem \eqref{step2} for $G$ is obtained as
\begin{align}
G^{t+1} &= \argmin_{G} \psi_{\Gamma_2}(G) + \frac{\rho}{2}\| G- (\GC^{t+1} + Y_2^t) \|_F^2 \\
&= \PC_{\Gamma_2}(\GC^{t+1} + Y_2^t),
\end{align}
where $\PC_{\Gamma_2}$ is a projection onto $\Gamma_2$.
Similarly, for $m$, we obtain an optimal solution as follows.
\begin{align}
m^{t+1} &= \argmin_{m} \psi_{\Gamma_1}(m) + \frac{\rho}{2}\| m- (\mu^{t+1} + Y_3^t) \|_F^2 \\
&= \PC_{\Gamma_1}(\mu^{t+1} + Y_3^t),
\end{align}
where $\PC_{\Gamma_1}$ is a projection onto $\Gamma_1$.

Finally, all  the steps are summarized in Algorithm \ref{Algo1}. 
\begin{algorithm} [H]  
\caption{Algorithm for solving \eqref{AP2}}
\begin{algorithmic}  \label{Algo1}
\REQUIRE Given data $Z^K \in \R^{n_1\times n_2 \times (K+p)}$, $\rho$, $\tau$,$a_1,a_2,b_1,b_2$
\ENSURE Matrix $\hm$ and tensor $\hg$
\STATE Initialize $\mu^{(0)}$, $\GC^{(0)}$,$R^{(0)}$, $m^{(0)}$, $G^{(0)}$, $Y_1^{(0)}$,$Y_2^{(0)}$,$Y_3^{(0)}$, and set $t=1$.
 \REPEAT 
 \STATE Update $\mu^{t+1}$ and $\GC^{t+1}$ by the following steps:
 \WHILE{not converge}
 \STATE Update $\mu,\GC$ by the equations \eqref{MM1} and \eqref{MM2}.
 \ENDWHILE
\STATE Update $R^{t+1}$, $G^{t+1}$, $m^{t+1}$ by solving \eqref{step2}
\STATE Update  
\vspace{-0.1in}
\begin{align*}
Y_1^{t+1} &= Y_1^t+ (\GC^{t+1}-R^{t+1}) \\Y_2^{t+1} &= Y_2^t+ (\GC^{t+1}-G^{t+1})\\ Y_3^{t+1} &= Y_3^t+ (\mu^{t+1}-m^{t+1})
\end{align*}
\vspace{-0.3in}
\STATE $t=t+1.$
 \UNTIL{Termination criterion is met.}
\end{algorithmic}
\end{algorithm}

\section{Numerical examples}\label{SecExper}
\subsection{Synthetic data}

We first experiment with Algorithm \ref{Algo1}  on synthetic data to see the performance of our method.
We generate the true $\GC\in \mathbb{R}^{(2n_1-1)\times (2n_2-1)\times p}$ with multi-rank $(r_1,r_2,\ldots,r_p)=(1,1,\ldots,1)$ by
$
	\GC_{ijk} = u^{(1)}_{i}u^{(2)}_{j}u^{(3)}_{k},
$
where $u^{(1)}_{i},u^{(2)}_{j}$, and $u^{(3)}_{k}$ are from uniform distribution $U(0,1)$. 
By our discrete approximation to Hawkes processes with memory depth $p$, we use a non-increasing function of $k$ for the $k$th frontal slice $\GC^{(k)}$ for $k=1,\ldots,p$. We also generate $\mu$ randomly from $U(0,1)$. 
The $\mu$ and $\GC$ are rescaled for a well-defined point process.
With the true $\mu$, $\GC$, we generate the synthetic data by
\begin{align} 
Z_{ijk} | \HC_{(k-1)}\sim {\rm Poisson}\left(\Delta\lambda_{ijk} \right)
 \end{align}
for $i \in \llbracket n_1 \rrbracket, j \in \llbracket n_2 \rrbracket$ and $k \in \llbracket K \rrbracket $ with given initial data $Z_{1-p}^0$. 
The initialization $\mu^{(0)}$ and $\GC^{(0)}$ are randomly generated with similar scales to their true values.  
To ensure that  the error terms  in different cases are at the same scale,  we use the relative error
	\begin{align}
	{\rm Merr} := \frac{\|\mu-\widehat \mu\|_F}{\|\mu\|_F} \ \text{ and }\  {\rm Gerr} :=\frac{\| \GC-\widehat \GC \|_F}{\|\GC\|_F}
	\end{align}
to evaluate the estimation of $\mu$ and $\GC$, respectively. 

\begin{figure*}[t]
\vskip 0.2in
\begin{center}
\centerline{
\subfigure{\includegraphics[height=6cm]{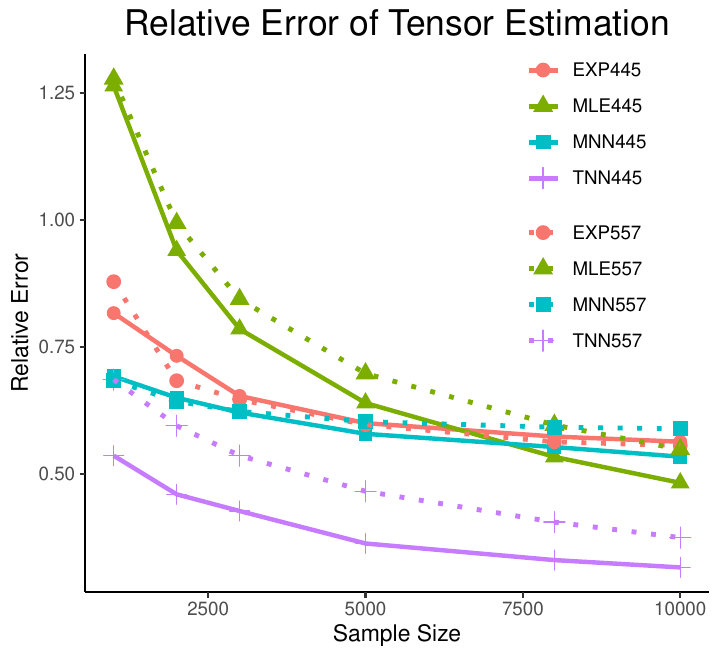}}
\hspace{1.6cm}
\subfigure{\includegraphics[height=6cm]{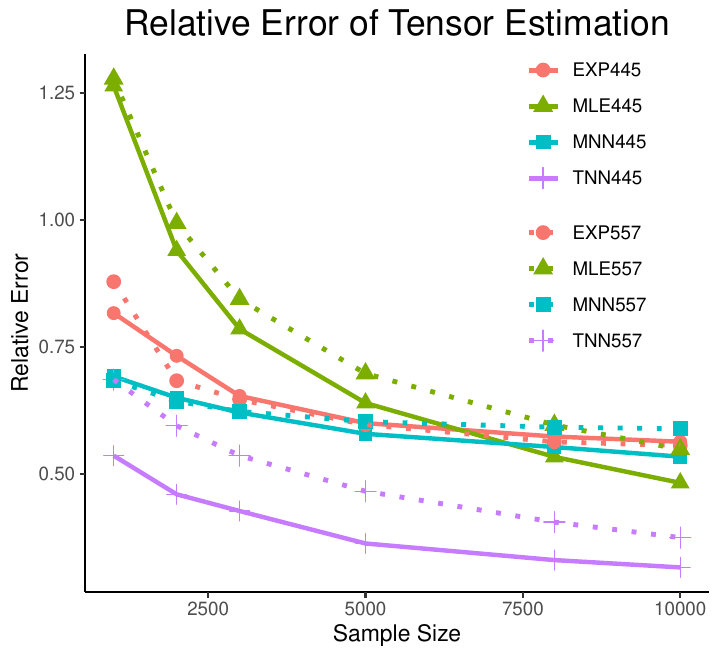}}
}
\caption{The relative errors of the estimated tensor kernel (left) and base intensity matrix (right) under different $n_1,n_2,p$, and sample size $K$. }
\label{Fig_TNNMLE}
\end{center}
\vskip -0.2in
\end{figure*}

We test the performance of our method with different $n_1,n_2$ and $p$, and compare it with the model without the low-rank constraint on $\GC$, and 
a widely used Hawkes process model with an exponential temporal decay function (e.g., \cite{zhou2013learning}).
We denote the proposed method  as ``TNN'', the maximum likelihood estimation method without the low-rank constraint as ``MLE", the estimation method with the exponential decay function (i.e., $ \alpha e^{-\alpha k}$, where $k = 1,\ldots p$ and $\alpha > 0 $ is a decay parameter.) as ``EXP",  and the estimation method with the matrix nuclear norm and the exponential decay function as ``MNN".

The parameters for each method are chosen based on the dataset that is disjoint from the training dataset. 
For parameter $\rho$ in Algorithm \ref{Algo1}, it is set to 0.002 for TNN, 0.001 for MLE, 0.06 for EXP and MNN.
Since $\rho$ serves as a dual step-size in ADMM, the performance of methods is robust to the mild change of $\rho$. For hyper-parameter $\tau$ on the regularized terms, a cross-validation-like method is exploited to tune the parameter and
it is set to 0.5 for TNN and 0.1 for MNN.
Two cases of experiments are carried out, and the representative results are shown in Figure \ref{Fig_TNNMLE}. The results of each case are averaged over five runs. Table \ref{average} in Appendix B shows the detailed estimation errors for each case.

In the experiments, we observe that
both the relative error of the matrix estimation and the tensor estimation decrease as the sample number grows and becomes close to zero.  
We notice that the numerical results correspond to our theoretical result on the upper bound of the estimation error mentioned in Remark \ref{convergeR}. 
Moreover, TNN outperforms MLE, EXP, and MNN. 
It shows the  computational efficiency of implementing the Fourier-transformed TNN
 when the tensor kernel has a low-rank structure, which commonly occurs in the real world 
 due to high correlations between locations and time. 
 It can be observed that TNN has advantages over EXP and MNN  
 for data with unknown or non-exponential temporal decay functions. 
 Moreover, TNN can capture more general spatio-temporal correlations than MNN, where the kernel is decomposed into a matrix and a temporal decay function. 
 We note that our model, TNN, can be easily applied to a wide range of discrete data.

 \subsection{Real data}
 We next apply our method to real-world data, the crime dataset in Atlanta, USA. 
 The dataset contains 47,245 burglary incidents in Atlanta from January 1, 2015, to February 28, 2017.
 The events  in the region where the latitude is from 33.71 to 33.76 and the longitude are from -84.43 to -84.38
 are considered. In the region, 9937 burglary incidents occurred during 789 days. We discretize the area
  into $5 \times 5 $ discrete space and the time with a 4-hour interval unit. 
 
 We use $p=5$ as a memory depth for the data, and the parameter $\tau$ is set to 3.5 for TNN and 0.4 for MNN.
The model is trained and tested with 80\% and 20\% of the data sequence, respectively.  
 
 To evaluate our model, two metrics are employed: First,  the metric FRQ, defined as the sum of the absolute difference in frequency of events between the predicted data and the true test data, is used. It is the frequency difference of burglary incidents in 25 subregions.  Second, the metric NLR, the sum of the negative log-likelihood function, is compared.
  
 Table \ref{table:1} shows
  numerical results on TNN, MLE, EXP and MNN. The numbers in the columns of FRQ (1) and FRQ (60) represent one instance of FRQ and 
 the average FRQ over 60 runs, respectively.
 In all metrics, TNN provides better results than MLE, EXP and MNN. 
 We observe again the clear advantage of exploiting a low-rank structure of the tensor kernel. 
 We note that TNN is implemented without any predefined decay parameter, whereas it is necessary for EXP and MNN.

 \begin{table}[ht]
\caption{ Evaluation of TNN, MLE, EXP and MNN with the Atlanta crime dataset. }
\label{table:1}
\vskip 0.05in
\begin{center}
\begin{sc}
\begin{tabular}{|c||c|c|c|}
	\hline
	 &\ \ \ FRQ (1) \ \ \ & \ \ \ FRQ (60) \ \ \ &\ \ \  \  \ NLR \  \ \ \  \   \\
	\hline
	\ \ \ TNN \ \ \ & 0.3474  &0.3677 & 4126.3   \\
	\hline
	MLE& 0.4456 &0.3732 &  4197.3  \\
	\hline
	EXP& 0.4509 &0.3790 &  4138.0  \\
	\hline
	MNN& 0.3823  &0.3681 & 4129.2 \\
	\hline 
	\end{tabular}
\end{sc}
\end{center}
\vskip -0.1in
\end{table}

\section{Conclusion}
We have studied the recovery of the base intensity matrix and the tensor of the discretized version of the kernel function for spatio-temporal Hawkes processes.
Using TNN, a formulation of the maximum likelihood estimation with the constraints has been proposed. Specifically, a precise theoretical upper bound for the sum of square errors of the proposed estimators has been presented. We have also applied the ADMM and MM algorithms to solve the proposed convex optimization problem. The numerical experiments demonstrate the efficiency of our method and support the theoretical results. 
For future work, non-convex optimization techniques will be investigated to estimate the matrix and the tensor kernel in the problem. It will be interesting to study whether the convex relaxation gap can be estimated and reduced by employing non-convex optimization methods.

\section*{Acknowledgments}
 This work was conducted while the first author was a master's student at H. Milton Stewart School of Industrial and Systems Engineering,  Georgia Institute of Technology.
 The work of Yao Xie is supported by NSF CAREER CCF-1650913, CMMI-2015787, DMS-1938106, and DMS-1830210. 


%

\section{Appendix}
\subsection{Proof of Theorem 3} 
We prove Theorem 3. For the simplicity of analysis, we let $\eta := (\mu,\GC)$.
Then, the problem is expressed as:
\begin{align} \label{MainP} 
&\min F(\eta)\! :=\! \sum_{k=1}^{K}\!\sum_{i=1}^{n_1}\!\sum_{j=1}^{n_2} \left[ \Delta \lambda_{ijk}(\eta) -  Z_{ijk}\!\ln(\Delta\lambda_{ijk}(\eta)) \right] \\
&\mbox{subject to} ~~ \ \eta \in \DC := \left\{ (\mu ,\GC) | \ a_1 \!\leq \mu_{ij}\! \leq b_1, \ a_2\! \leq \GC_{ijk}\! \leq b_2, \right. \\
 & \left.  \qquad \qquad  \|\GC\|_{\rm TNN} \leq b_2\sqrt{\gamma(2n_1-1)(2n_2-1)p} \ \right\},
\end{align}
where $\lambda, Z \in \R^{n_1\times n_2\times K}$.

We now define the KL-divergence between two Poisson distributions. For any two Poisson mean $p$ and $q$, the KL  divergence is defined as follows:
\begin{align}
D(p||q ) := p\ln(p/q)-(p-q).
\end{align}
Similarly, the Hellinger distance  for Poisson distributions is defined:
\begin{align}
H^2(p||q ) := 2-2\exp\left\{-\frac{1}{2}(\sqrt{p} -\sqrt{q})^2 \right\}.
\end{align}

\noindent
 Let $\theta$ be a true parameter that we aim to estimate. For any $ \eta \in \DC$, we have 
\begin{align}
& \quad F(\eta) - F(\theta) \\
&=  \sum_{k=1}^{K} F_k(\eta) - F_k (\theta) \\
&=  \sum_{k=1}^{K} \left\{ F_k(\eta) +E_{k-1}[F_k(\eta)]  - E_{k-1}[F_k(\eta)]- F_k (\theta)\right. \\
&\qquad\qquad\left. + E_{k-1}[F_k(\theta)] - E_{k-1}[F_k(\theta)]  \right\} \\
&= \sum_{k=1}^{K} \left\{ E_{k-1}[F_k(\eta)-F_k(\theta) ] +  F_k(\eta) - E_{k-1}[F_k(\eta)]\right. \\
&\qquad\qquad\left.- F_k (\theta) +E_{k-1}[F_k(\theta)]   \right\}, \label{maineq}
\end{align}
where $F_k(\eta) := \sum_{i=1}^{n_1}\sum_{j=1}^{n_2} ( \Delta\lambda_{ijk}(\eta) -  Z_{ijk}\ln(\Delta\lambda_{ijk}(\eta)))$ and
$E_{k-1}$ denotes the conditional expectation taken with respect to $Z^{k}$ given $\HC_{k-1}$.

Observe that 
\begin{align} \label{Term1}
&\quad \sum_{k=1}^{K}  E_{k-1}[F_k(\eta)-F_k(\theta) ] \\
& =  \sum_{k=1}^{K}\sum_{i=1}^{n_1}\sum_{j=1}^{n_2}  \Delta\lambda_{ijk}(\theta) \ln\frac{\lambda_{ijk}(\theta)}{\lambda_{ijk}(\eta )} - \Delta( \lambda_{ijk}(\theta) -\lambda_{ijk}(\eta ))\\
&= \sum_{k=1}^{K}\sum_{i=1}^{n_1}\sum_{j=1}^{n_2} \Delta D(\lambda_{ijk}(\theta) || \lambda_{ijk}(\eta)).
\end{align}
Since our estimator $\widehat \theta$ is the optimal solution to the problem \eqref{MainP}, we obtain that $F(\widehat \theta) - F(\theta) \leq 0$. From \eqref{maineq} and \eqref{Term1}, we have

\begin{align}
&\quad \sum_{k=1}^{K}\sum_{i=1}^{n_1}\sum_{j=1}^{n_2}  \Delta D(\lambda_{ijk}(\theta) || \lambda_{ijk}(\widehat\theta)) \\
&  \leq \sum_{k=1}^{K} - F_k (\widehat \theta) + E_{k-1}[F_k(\widehat \theta)]  + F_k (\theta) - E_{k-1}[F_k(\theta)]    \\
&= \!\sum_{k=1}^{K}\!\sum_{i=1}^{n_1}\!\sum_{j=1}^{n_2}  \{  - (\Delta \lambda_{ijk}( \theta) -Z_{ijk}  )\ln(\Delta \lambda_{ijk}(\widehat \theta)) \\
 &\qquad\qquad + ( \Delta \lambda_{ijk}( \theta) -Z_{ijk} ) \ln(\Delta \lambda_{ijk}(\theta)) \}   \label{Noise}  \\ 
&= \sum_{k=1}^{K}\sum_{i=1}^{n_1}\sum_{j=1}^{n_2}  (\Delta \lambda_{ijk}(\theta) -  Z_{ijk})\ln\left( \frac{ \lambda_{ijk}( \theta)}{ \lambda_{ijk}(\widehat \theta) }\right ) . 
\end{align}
We first derive the lower bound and then the upper bound for the inequality \eqref{Noise}.

\subsection*{Lower bound for KL-divergence}
For a fixed $Z^K$, we describe how to obtain the lower bound for 
\begin{align}
\sum_{k=1}^{K}\sum_{i=1}^{n_1}\sum_{j=1}^{n_2}  &\Delta D(\lambda_{ijk}(\theta) || \lambda_{ijk}(\widehat\theta)). 
\end{align}
From the information theory, we know that
\begin{align}
 D(\lambda_{ijk}(\theta) || \lambda_{ijk}(\widehat\theta))  \geq H^2(\lambda_{ijk}(\theta) || \lambda_{ijk}(\widehat\theta)). 
 \end{align}
 
To  obtain the lower and upper bound for any $ \lambda_{ijk}(\theta)$ with $Z^K$, we define
 the mapping  $W^{i j }(\cdot) :\R^{n_1\times n_2 \times p} \rightarrow \R^{(2n_1-1)\times (2n_2-1) \times p}$ as follows:
\begin{align}
&(W^{i j }(Z^{k-1}_{k-p}))_{\ti \tj \tk} \overset{\vartriangle}{=} \begin{cases}
  Z_{n_1-(\ti-i),n_2-(\tj-j),k-\tk},  &\quad \text{if } 
  \begin{aligned}[t]
  i &\leq \ti \leq i + n_1 -1, \\   j &\leq \tj \leq j +n_2-1, \\
  1 &\leq \tk \leq p \\
    \end{aligned}
  \\
    0,  &\quad\text{ otherwise. } 
  \end{cases} 
\end{align}
Then, we  can express $\lambda_{ijk}(\theta)$ as
\begin{align}
\lambda_{ijk}(\theta) = \mu_{ij} + \la W^{i j }(Z^{k-1}_{k-p}) ,\GC \ra,
\end{align}
where $\la\cdot,\cdot \ra$ is an inner product for tensors.

Let $\E \in \R^{(2n_1-1)\times (2n_2-1) \times p}$ be a tensor of all ones. 
We define 
\begin{align}
l := \min_{k} \{ \la \E , W^{i j }(Z^{k-1}_{k-p}) \ra \} = \min_{k} \{ \| Z^{k-1}_{k-p}\|_1 \}
\end{align}
and
\begin{align}
u := \max_{1 \leq k \leq K} \{ \| W^{i j }(Z^{k-1}_{k-p})\|_{\rm spec} \}= \max_{1 \leq k \leq K} \{ \| Z^{k-1}_{k-p}\|_{\rm spec} \}.
\end{align}
For any $ \lambda_{ijk}(\theta| \HC_{k-1})$, the lower bound is
\begin{align}
 \lambda_{ijk}(\theta) \geq a_1 + a_2 \la \E , W^{i j }(Z^{k-1}_{k-p}) \ra \geq a_1+a_2l,
\end{align}
and the upper bound is 
\begin{align}
 \lambda_{ijk}(\theta) &\leq b_1 +  \la \GC , W^{i j }(Z^{k-1}_{k-p}) \ra \\
 &\leq b_1 +\| W^{i j }(Z^{k-1}_{k-p})\|_{\rm spec}\|\GC\|_{\rm TNN}\\
&\leq b_1+ub_2\sqrt{\gamma(2n_1-1)(2n_2-1)p}
\end{align}
by Cauchy-Schwartz inequality and the assumptions. As a result,  given $Z^K$, 
\begin{align}
 \underbar{$J$} &:= a_1+a_2l \leq \lambda_{ijk}(\theta) \\
 &\leq b_1+ub_2\sqrt{\gamma(2n_1-1)(2n_2-1)p}:= \bar J,
 \forall i,j,k, \forall \theta \in \DC.
 \end{align}
By Lemma 8 in \cite{cao2016poisson}, for all $T \geq \frac{1}{2}\left(\sqrt{\lambda_{ijk}(\theta)} -\sqrt{\lambda_{ijk}(\widehat \theta)}\right)^2$,
 it holds that 
\begin{align}
H^2(\lambda_{ijk}(\theta) || \lambda_{ijk}(\widehat \theta)) \geq \frac{1-e^{-T}}{4\bar J T} [ \lambda_{ijk}(\theta)- \lambda_{ijk}(\widehat \theta) ]^2.
\end{align}
Taking $T = 2\bar J$, we have 
\begin{align}
H^2(\lambda_{ijk}(\theta) || \lambda_{ijk}(\widehat \theta)) \geq \frac{1-e^{-2\bar J}}{8\bar J^2 } [ \lambda_{ijk}(\theta)- \lambda_{ijk}(\widehat \theta) ]^2.
\end{align}

For the subsequent discussion, we need the following notation. Let $E_{ij} \in \R^{n_1 \times n_2}$ be a matrix whose 
$ij$th entry is one and all the other entries are zero. We also denote the number of parameters as $d := n_1n_2 +(2n_1-1)(2n_2-1)p$. Let
\begin{align}
\beta &:= [\textsf{vec} (\mu) ; \textsf{vec}(\GC) ] \in \R^d, \\
c_{ij}(Z^{k-1}_{k-p}) &:= [\textsf{vec}(E_{ij}) ; \textsf{vec}(W^{i j }(Z^{k-1}_{k-p})) ] \in \R^d, \\
A_{ij}(Z^{k-1}_{k-p}) &:= c_{ij}(Z^{k-1}_{k-p}) c_{ij}(Z^{k-1})_{k-p} ^\top \in \R^{d \times d}, \\
 A[Z^K] &:= \frac{1}{K} \sum_{k=1}^{K}\sum_{i=1}^{n_1}\sum_{j=1}^{n_2} A_{ij}(Z^{k-1}_{k-p}) \in \R^{d \times d},
\end{align}
where \textsf{vec} is a vectorization operator.

Then, we can represent $\lambda_{ijk}(\theta)$ as follows.
\begin{align}
\lambda_{ijk}(\theta) = c_{ij}(Z^{k-1}_{k-p})^\top \beta.
\end{align}
Thus, 
\begin{align}
& H^2(\lambda_{ijk}(\theta) || \lambda_{ijk}(\widehat \theta))\\
 \geq &  \frac{1-e^{-2\bar J}}{8\bar J^2} [ \lambda_{ijk}(\theta)- \lambda_{ijk}(\widehat \theta) ]^2\\
 = & \frac{1-e^{-2\bar J}}{8\bar J^2} [  c_{ij}(Z^{k-1}_{k-p})^\top \beta - c_{ij}(Z^{k-1}_{k-p})^\top \widehat\beta]^2 \\
= & \frac{1-e^{-2\bar J}}{8\bar J^2} (\beta-\widehat\beta)^\top ( c_{ij}(Z^{k-1}_{k-p}) c_{ij}(Z^{k-1}_{k-p}) ^\top ) (\beta-\widehat\beta) \\
 = & \frac{1-e^{-2\bar J}}{8\bar J^2} (\beta-\widehat\beta)^\top A_{ij}(Z^{k-1}_{k-p}) (\beta-\widehat\beta). 
\end{align}
Note that for given data $Z^{k-1}_{k-p}$, $A_{ij}(Z^{k-1}_{k-p})$ is positive semidefinite ($A_{ij}(Z^{k-1}_{k-p}) \succeq 0$).
We use the condition number in Definition 2 to obtain the lower bound for \eqref{Noise}:
\begin{equation}\label{lb}
\begin{aligned} 
&\sum_{k=1}^{K}\sum_{i=1}^{n_1}\sum_{j=1}^{n_2} \Delta D(\lambda_{ijk}(\theta) || \lambda_{ijk}(\widehat\theta )) \\
\geq& \sum_{k=1}^{K}\sum_{i=1}^{n_1}\sum_{j=1}^{n_2}\Delta  H^2(\lambda_{ijk}(\theta) || \lambda_{ijk}(\widehat \theta)) \\
\geq& \Delta K\!\sum_{k=1}^{K}\!\sum_{i=1}^{n_1}\!\sum_{j=1}^{n_2}\! \frac{1-e^{-2\bar J}}{8\bar J^2 K} (\beta-\widehat\beta)^\top A_{ij}(Z^{k-1}_{k-p}) (\beta-\widehat\beta) \\
\geq&  \Delta K\frac{1-e^{-2\bar J}}{8\bar J^2 } \delta_{2}[A[Z^K]] \|\beta -\widehat\beta \|_2^2 \\
=& \Delta K\frac{1-e^{-2\bar J}}{8\bar J^2} \delta_{2}[A[Z^K]]  ( \| \mu-\hm \|_F^2 + \| \GC-\hg \|_F^2 ).
\end{aligned}
\end{equation}

The last inequality follows from  Definition 2.

\subsection*{Upper bounds for the random term}
 We next derive the upper bound on \eqref{Noise}.
The upper bound can be written as 

\begin{align}
& \sum_{k=1}^{K}\sum_{i=1}^{n_1}\sum_{j=1}^{n_2}  (\Delta \lambda_{ijk}(\theta) -  Z_{ijk})\ln\left( \frac{ \lambda_{ijk}( \theta)}{ \lambda_{ijk}(\widehat \theta) }\right ) \\
&=  \sum_{k=1}^{K} \left \la E , ( \Delta \lambda^{(k)}(\theta) - Z^{(k)}) \circ \ln\left( \frac{ \lambda^{(k)}( \theta)}{ \lambda^{(k)}(\widehat \theta) }\right ) \right \ra   \\
&  \leq \sup_{\eta \in \DC} \sum_{k=1}^{K} \left \la E , ( \Delta \lambda^{(k)}(\theta) - Z^{(k)}) \circ \ln\left( \frac{ \lambda^{(k)}( \theta)}{ \lambda^{(k)}(\eta) }\right ) \right \ra, \label{UB}
\end{align} 
where $E \in \R^{n_1 \times n_2}$ is a matrix of all ones, $\lambda^{(k)} , Z^{(k)}$ are $k$th frontal slice of tensor $\lambda$ and $Z$, respectively,
$\circ$ is the Hadamard product, and $\ln\left( \frac{ \lambda^{(k)}( \theta)}{ \lambda^{(k)}(\widehat \theta) }\right ) \in \R^{n_1\times n_2}$ is a matrix whose $ij$th entry is equal to $\ln\left( \frac{ \lambda_{ijk}( \theta)}{ \lambda_{ijk}(\widehat \theta) }\right )$.  
For the analysis of \eqref{UB}, we define 
\begin{align}
\xi_k :=  \textsf{vec} \left [( \Delta  \lambda^{(k)}(\theta) - Z^{(k)} ) \circ \ln\left( \frac{ \lambda^{(k)}( \theta)}{ \lambda^{(k)}(\eta) }\right ) \right ] \in \R^{n_1n_2}.
\end{align}
Note that $\xi_k$ is a \emph{martingale difference} vector.

In the subsequent discussion, 
we will apply the Azuma-Hoeffding inequality and union bound property to derive an upper bound. 
We need the condition, $| (\xi_k)_s | \leq b_s $ for all $s= 1, \cdots ,n_1n_2$, to apply the Azuma-Hoeffding inequality.
The bounds can be obtained by applying the following Poisson concentration inequality.
\begin{lemma}\label{PoiIne}
For $Y \sim Pois(\lambda)$, for all $t>0$, it holds that 
\begin{align}
\mathbb{P} \{ | Y - \lambda | \geq t \} \leq 2e^{-\frac{t^2}{2(\lambda+t)} }.
\end{align}
\end{lemma}
By Lemma \ref{PoiIne}, for $\eps >0$,
\begin{align}
\mathbb{P}  \{ | \Delta \lambda_{ijk}(\theta) - &  Z_{ijk}) | \geq \eps |  \HC_{k-1} \} \leq 2e^{-\frac{\eps^2}{2(\lambda_{ijk}(\theta)+\eps)} } \leq  2e^{-\frac{\eps^2}{2(\Delta \bar J +\eps)} }.
\end{align}
By the tower property for conditional expectations,
\begin{align}
&\E\left[\mathbb{P} \{ |\Delta  \lambda_{ijk}(\theta) -  Z_{ijk}) | \geq \eps | \HC_{k-1} \} \right]\\
=&\mathbb{P} \{ | \Delta \lambda_{ijk}(\theta) -  Z_{ijk} | \geq \eps \} 
\leq 2e^{-\frac{\eps^2}{2(\Delta \bar J +\eps)} } .
\end{align}
Therefore, by applying the union bound property, 
\begin{align}
| \Delta \lambda_{ijk}(\theta) -  Z_{ijk} | \leq \eps, \forall i,j,k,
\end{align}
with probability $1- 2n_1n_2Ke^{-\frac{\eps^2}{2(\Delta \bar J +\eps)} } $.
 Since $\theta, \eta \in \DC$, we have 
\begin{align}
\left | ( \Delta \lambda_{ijk}(\theta) -  Z_{ijk} )\ln\left( \frac{ \lambda_{ijk}( \theta)}{  \lambda_{ijk}(\eta)} \right ) \right| \leq \eps  \ln \frac{ \bar J}{\underbar{$J$} }   := \eps' , 
\end{align}
with probability $1- 2n_1n_2Ke^{-\frac{\eps^2}{2(\Delta \bar J +\eps)} } $.  Note that this shows each entry in $\xi_k$ is not upper bounded by $\eps'$ with a small probability.

Now we apply the following Theorem. 
\begin{theorem} 
[\upshape{Theorem 32,33 in \cite{chung2006concentration}}]\label{martg}
Consider a random variable $X$ and a filtration $\{\FC_0, \ldots \FC_n\}$.
Suppose $X_0, X_1, \ldots X_n$ is a martingale sequence such that $X_i = \E[X|\FC_i]$. For $t>0$, it holds that 
\[
    \P(| X- \E X| \geq t) \leq 2e^{-\frac{t^2}{2\sum_ic_i^2}} + \sum_i \P(|X_i-X_{i-1}| \geq c_i),
\]
where $c_1, \ldots, c_n$ are non-negative values. 
\end{theorem}
 For fixed $s$, we define ``bad events'' as a set such that $|(\xi_k)_s| > \eps'$ for any $k = 1 \ldots K$. 
By Theorem \ref{martg}, the generalized Azuma- Hoeffding inequality can be applied to the sum of unbounded martingale difference with a probability of the ``bad events". 

For $t>0$, we obtain 
\begin{align}
\P \left\{ \Bigg| \left(\sum_{k=1}^{K} \xi_k \right)_s \Bigg| \geq t \right\} \leq 2e^{-\frac{t^2}{2K\eps'^2}} + \P(\text{``bad events"}) 
\end{align}
and it implies that for $x> 0$,
\begin{align}
\P\left\{ \Bigg| \left(\sum_{k=1}^{K} \xi_k \right)_s \Bigg| \geq \sqrt{2\eps'^2xK} \right\} \leq 2e^{-x} + 2Ke^{-\frac{\eps^2}{2(\Delta \bar J +\eps)}}  .
\end{align}
By the union bound, we have 
\begin{align}
 \left\|\sum_{k=1}^{K} \xi_k  \right\|_{\infty} \leq \sqrt{2\eps'^2 xK}
\end{align}
with probability $1- 2n_1n_2Ke^{-\frac{\eps^2}{2(\bar J +\eps)} } -2n_1n_2e^{-x}$.

Let $\alpha_1=n_1n_2Ke^{-\frac{\eps^2}{2(\Delta \bar J +\eps)} }$ and $\alpha_2=n_1n_2e^{-x}$, where $\bar J=b_1+u\sqrt{\gamma(2n_1-1)(2n_2-1)p}$ and $x>0$.
By simple computation,  we have 
\begin{align}
\eps  = &\ln\frac{n_1n_2K}{\alpha_1} + \sqrt{\ln^2\frac{n_1n_2K}{\alpha_1} + 2\Delta \bar J\ln\frac{n_1n_2K}{\alpha_1}} \\
\leq & \ \max \left \{ 2\sqrt{\Delta \bar J\ln\frac{n_1n_2K}{\alpha_1} } , \  4 \ln\frac{n_1n_2K}{\alpha_1}    \right \}.
\end{align} 
Hence, it follows that  
\begin{align}
 \left\|\sum_{k=1}^{K}\! \xi_k \!\right\|_{\infty} \leq & \sqrt{2K} \ln \frac{ \bar J}{\underbar{$J$} } \sqrt{\ln\frac{n_1n_2}{\alpha_2}} \cdot \max \left \{ 2\sqrt{\Delta \bar J\ln\frac{n_1n_2K}{\alpha_1} } , \  4 \ln\frac{n_1n_2K}{\alpha_1}    \right \}
\end{align}
with probability $1-2\alpha_1 -2\alpha_2$.
Finally, the upper bound is
\begin{align}
& \sum_{k=1}^{K}\sum_{i=1}^{n_1}\sum_{j=1}^{n_2} \Delta  D(\lambda_{ijk}(\theta) || \lambda_{ijk}(\widehat \theta)) \\
&\leq \sup_{\eta \in \DC} \sum_{k=1}^{K} \left \la E , ( \Delta \lambda^{(k)}(\theta) - Z^{(k)}) \circ \ln\left( \frac{ \lambda^{(k)}( \theta)}{ \lambda^{(k)}(\eta) }\right ) \right \ra  \\
&\leq   \|\textsf{vec}(E)\|_1  \|\sum_{k=1}^{K} \xi_k \|_{\infty}  \\
&\leq n_1n_2 \sqrt{2K} \ln \frac{ \bar J}{\underbar{$J$} } \sqrt{\ln\frac{n_1n_2}{\alpha_2}}  \cdot \max \left \{ 2\sqrt{\Delta \bar J\ln\frac{n_1n_2K}{\alpha_1} } , \  4 \ln\frac{n_1n_2K}{\alpha_1}    \right \} \label{ub}
\end{align}
with  probability at least $1-2\alpha_1 -2\alpha_2 $.
We obtain Theorem 3 by combining  \eqref{lb} and \eqref{ub}.

\subsection{Numerical results for simulation}

Table \ref{average} demonstrates the results for the simulations in Section 5. ``TNN'' denotes our method, which involves low-rank constraints using Fourier transformed nuclear norm, while ``MLE'' denotes the maximum likelihood method without such constraint, ``EXP'' denotes the estimation method with fixed exponential temporary decay function, and ``MNN" denotes the estimation method with the matrix nuclear norm and the exponential decay function.
For each case and each sample size, the experiment was repeated five times for each method. The visualization is presented in Figure 2 in Section 5.

\begin{table*}[htbp]
\captionsetup{size=footnotesize}
\caption{The numerical estimation error with different $n_1,n_2, p$ and $K$}  \label{average}
\setlength\tabcolsep{0pt} 
\footnotesize\centering
\smallskip 
\begin{tabular*}{\textwidth}{@{\extracolsep{\fill}}ccccccccc}
\toprule
Case & Method &Error &$K$=1000 & $K$=2000 & $K$=3000 & $K$=5000 & $K$=8000 & $K$=10000   \\
\midrule
\multirow{4}{*}{$n_1=4,\ n_2=4,\ p=5$}&
	\multirow{2}{*}{TNN}& Gerr & 0.536 & 0.460 & 0.427 &0.363 &0.330 &0.316 \\
	&& Merr & 0.202 & 0.131 & 0.094 &0.080 &0.064 &0.061 \\ 
	\cline{2-9}
	&\multirow{2}{*}{MLE}& Gerr & 1.264 & 0.940 & 0.785 &0.640 &0.534 &0.482 \\
	&& Merr & 0.202 & 0.144 & 0.110 &0.094 &0.066 &0.063 \\
	\cline{2-9}
	&\multirow{2}{*}{EXP}& Gerr & 0.817 & 0.732 & 0.654 &0.600 &0.574 &0.563 \\
	&& Merr & 0.296 & 0.174 & 0.133 &0.112 &0.103 &0.109 \\
	\cline{2-9}
	&\multirow{2}{*}{MNN}& Gerr & 0.692 & 0.650 & 0.621 &0.597 &0.553 &0.534 \\
	&& Merr & 0.215 & 0.163 & 0.143 &0.148 &0.144 &0.137 \\
	\hline
	
	\multirow{4}{*}{$n_1=5,\ n_2=5,\ p=7$}&
	\multirow{2}{*}{TNN}& Gerr & 0.686 & 0.595 & 0.536 &0.466 &0.405 &0.375\\
	&&Merr & 0.241 & 0.178 & 0.192 & 0.133 &0.116 &0.105 \\
	\cline{2-9}
	&\multirow{2}{*}{MLE}& Gerr & 1.278 & 0.994 & 0.844 &0.698 &0.596 &0.548 \\
	&& Merr & 0.378 & 0.315 & 0.297 &0.237 &0.249 &0.241 \\
	\cline{2-9}
	&\multirow{2}{*}{EXP}& Gerr & 0.878 & 0.984 & 0.646 &0.597 &0.563 &0.556 \\
	&& Merr & 0.386 & 0.456 & 0.385 &0.312 &0.267 &0.239 \\
	\cline{2-9}
	&\multirow{2}{*}{MNN}& Gerr & 0.684 & 0.643 & 0.623 &0.603 &0.592 &0.589 \\
	&& Merr & 0.462 & 0.380 & 0.323 &0.256 &0.217 &0.200 \\
\bottomrule
\end{tabular*}
\end{table*}


\begin{thebibliography}{10}

\bibitem{bacry2020sparse}
Emmanuel Bacry, Martin Bompaire, St\'{e}phane Gax\''{i}ffas, and Jean-Francois
  Muzy.
\newblock Sparse and low-rank multivariate hawkes processes.
\newblock {\em Journal of Machine Learning Research}, 21(50):1--32, 2020.

\bibitem{bahadori2014fast}
Mohammad~Taha Bahadori, Qi~Rose Yu, and Yan Liu.
\newblock Fast multivariate spatio-temporal analysis via low rank tensor
  learning.
\newblock In {\em Advances in neural information processing systems}, pages
  3491--3499, 2014.

\bibitem{bengua2017efficient}
Johann~A Bengua, Ho~N Phien, Hoang~Duong Tuan, and Minh~N Do.
\newblock Efficient tensor completion for color image and video recovery:
  Low-rank tensor train.
\newblock {\em IEEE Transactions on Image Processing}, 26(5):2466--2479, 2017.

\bibitem{boyd_vandenberghe_2004}
Stephen Boyd and Lieven Vandenberghe.
\newblock {\em Convex Optimization}.
\newblock Cambridge University Press, 2004.

\bibitem{cai2019nonconvex}
Changxiao Cai, Gen Li, H~Vincent Poor, and Yuxin Chen.
\newblock Nonconvex low-rank tensor completion from noisy data.
\newblock In {\em Advances in Neural Information Processing Systems}, pages
  1863--1874, 2019.

\bibitem{candes2010matrix}
Emmanuel~J Candes and Yaniv Plan.
\newblock Matrix completion with noise.
\newblock {\em Proceedings of the IEEE}, 98(6):925--936, 2010.

\bibitem{candes2009exact}
Emmanuel~J Cand{\`e}s and Benjamin Recht.
\newblock Exact matrix completion via convex optimization.
\newblock {\em Foundations of Computational mathematics}, 9(6):717, 2009.

\bibitem{candes2010power}
Emmanuel~J Cand{\`e}s and Terence Tao.
\newblock The power of convex relaxation: Near-optimal matrix completion.
\newblock {\em IEEE Transactions on Information Theory}, 56(5):2053--2080,
  2010.

\bibitem{cao2016poisson}
Y.~{Cao} and Y.~{Xie}.
\newblock Poisson matrix recovery and completion.
\newblock {\em IEEE Transactions on Signal Processing}, 64(6):1609--1620, 2016.

\bibitem{cao2015poisson}
Yang Cao and Yao Xie.
\newblock Poisson matrix completion.
\newblock In {\em 2015 IEEE International Symposium on Information Theory
  (ISIT)}, pages 1841--1845. IEEE, 2015.

\bibitem{carroll1970analysis}
J~Douglas Carroll and Jih-Jie Chang.
\newblock Analysis of individual differences in multidimensional scaling via an
  n-way generalization of “eckart-young” decomposition.
\newblock {\em Psychometrika}, 35(3):283--319, 1970.

\bibitem{chung2006concentration}
Fan Chung and Linyuan Lu.
\newblock Concentration inequalities and martingale inequalities: a survey.
\newblock {\em Internet mathematics}, 3(1):79--127, 2006.

\bibitem{fazel2013hankel}
Maryam Fazel, Ting~Kei Pong, Defeng Sun, and Paul Tseng.
\newblock Hankel matrix rank minimization with applications to system
  identification and realization.
\newblock {\em SIAM Journal on Matrix Analysis and Applications},
  34(3):946--977, 2013.

\bibitem{gandy2011tensor}
Silvia Gandy, Benjamin Recht, and Isao Yamada.
\newblock Tensor completion and low-n-rank tensor recovery via convex
  optimization.
\newblock {\em Inverse Problems}, 27(2):025010, 2011.

\bibitem{goldfarb2014robust}
Donald Goldfarb and Zhiwei Qin.
\newblock Robust low-rank tensor recovery: Models and algorithms.
\newblock {\em SIAM Journal on Matrix Analysis and Applications},
  35(1):225--253, 2014.

\bibitem{guo2013accelerating}
Ce~Guo and Wayne Luk.
\newblock Accelerating maximum likelihood estimation for hawkes point
  processes.
\newblock In {\em 2013 23rd International Conference on Field programmable
  Logic and Applications}, pages 1--6. IEEE, 2013.

\bibitem{hardiman2013critical}
Stephen~J Hardiman, Nicolas Bercot, and Jean-Philippe Bouchaud.
\newblock Critical reflexivity in financial markets: a hawkes process analysis.
\newblock {\em The European Physical Journal B}, 86(10):442, 2013.

\bibitem{harshman1970foundations}
Richard~A Harshman et~al.
\newblock Foundations of the parafac procedure: Models and conditions for an"
  explanatory" multimodal factor analysis.
\newblock 1970.

\bibitem{hillar2013most}
Christopher~J Hillar and Lek-Heng Lim.
\newblock Most tensor problems are np-hard.
\newblock {\em Journal of the ACM (JACM)}, 60(6):1--39, 2013.

\bibitem{juditsky2020convex}
Anatoli Juditsky, Arkadi Nemirovski, Liyan Xie, and Yao Xie.
\newblock Convex recovery of marked spatio-temporal point processes.
\newblock {\em arXiv preprint arXiv:2003.12935}, 2020.

\bibitem{kilmer2011factorization}
Misha~E Kilmer and Carla~D Martin.
\newblock Factorization strategies for third-order tensors.
\newblock {\em Linear Algebra and its Applications}, 435(3):641--658, 2011.

\bibitem{INARhawkes}
Matthias Kirchner.
\newblock An estimation procedure for the hawkes process.
\newblock {\em Quantitative Finance}, 17(4):571--595, 2017.

\bibitem{kirchner2018nonparametric}
Matthias Kirchner and A~Bercher.
\newblock A nonparametric estimation procedure for the hawkes process:
  comparison with maximum likelihood estimation.
\newblock {\em Journal of Statistical Computation and Simulation},
  88(6):1106--1116, 2018.

\bibitem{kobayashi2016tideh}
Ryota Kobayashi and Renaud Lambiotte.
\newblock Tideh: Time-dependent hawkes process for predicting retweet dynamics.
\newblock {\em arXiv preprint arXiv:1603.09449}, 2016.

\bibitem{koltchinskii2011nuclear}
Vladimir Koltchinskii, Karim Lounici, Alexandre~B Tsybakov, et~al.
\newblock Nuclear-norm penalization and optimal rates for noisy low-rank matrix
  completion.
\newblock {\em The Annals of Statistics}, 39(5):2302--2329, 2011.

\bibitem{liu2012tensor}
Ji~Liu, Przemyslaw Musialski, Peter Wonka, and Jieping Ye.
\newblock Tensor completion for estimating missing values in visual data.
\newblock {\em IEEE transactions on pattern analysis and machine intelligence},
  35(1):208--220, 2012.

\bibitem{ogata1998space}
Yosihiko Ogata.
\newblock Space-time point-process models for earthquake occurrences.
\newblock {\em Annals of the Institute of Statistical Mathematics},
  50(2):379--402, 1998.

\bibitem{omi2017hawkes}
Takahiro Omi, Yoshito Hirata, and Kazuyuki Aihara.
\newblock Hawkes process model with a time-dependent background rate and its
  application to high-frequency financial data.
\newblock {\em Physical Review E}, 96(1):012303, 2017.

\bibitem{reynaud2010adaptive}
Patricia Reynaud-Bouret, Sophie Schbath, et~al.
\newblock Adaptive estimation for hawkes processes; application to genome
  analysis.
\newblock {\em The Annals of Statistics}, 38(5):2781--2822, 2010.

\bibitem{romera2013new}
Bernardino Romera-Paredes and Massimiliano Pontil.
\newblock A new convex relaxation for tensor completion.
\newblock In {\em Advances in Neural Information Processing Systems}, pages
  2967--2975, 2013.

\bibitem{semerci2014tensor}
Oguz Semerci, Ning Hao, Misha~E Kilmer, and Eric~L Miller.
\newblock Tensor-based formulation and nuclear norm regularization for
  multienergy computed tomography.
\newblock {\em IEEE Transactions on Image Processing}, 23(4):1678--1693, 2014.

\bibitem{tucker1966some}
Ledyard~R Tucker.
\newblock Some mathematical notes on three-mode factor analysis.
\newblock {\em Psychometrika}, 31(3):279--311, 1966.

\bibitem{zhang2016exact}
Zemin Zhang and Shuchin Aeron.
\newblock Exact tensor completion using t-svd.
\newblock {\em IEEE Transactions on Signal Processing}, 65(6):1511--1526, 2016.

\bibitem{zhang2014novel}
Zemin Zhang, Gregory Ely, Shuchin Aeron, Ning Hao, and Misha Kilmer.
\newblock Novel methods for multilinear data completion and de-noising based on
  tensor-svd.
\newblock In {\em Proceedings of the IEEE conference on computer vision and
  pattern recognition}, pages 3842--3849, 2014.

\bibitem{zhou2013learning}
Ke~Zhou, Hongyuan Zha, and Le~Song.
\newblock Learning social infectivity in sparse low-rank networks using
  multi-dimensional hawkes processes.
\newblock In {\em Artificial Intelligence and Statistics}, pages 641--649,
  2013.

\bibitem{zhuang2019semiparametric}
Jiancang Zhuang and Jorge Mateu.
\newblock A semiparametric spatiotemporal hawkes-type point process model with
  periodic background for crime data.
\newblock {\em Journal of the Royal Statistical Society: Series A (Statistics
  in Society)}, 182(3):919--942, 2019.

\end{thebibliography}
\end{document}